\documentclass[10pt,twocolumn,letterpaper]{article}

\usepackage[pagenumbers]{iccv}      

\def\pz{{\phantom{0}}}

\definecolor{iccvblue}{rgb}{0.21,0.49,0.74}
\definecolor{applegreen}{RGB}{100, 200, 0}
\usepackage[pagebackref,breaklinks,colorlinks,allcolors=iccvblue]{hyperref}
\usepackage{multirow}
\usepackage{array}
\usepackage[capitalize]{cleveref}
\usepackage{algorithm}
\usepackage{algorithmic}

\def\paperID{1181} 
\def\confName{ICCV}
\def\confYear{2025}

\title{Robust Latent Matters:\\ Boosting Image Generation with Sampling Error Synthesis}

\author{Kai Qiu$^{1*}$ \quad Xiang Li$^{1}$\thanks{Equal contribution.}\,\,\thanks{Now at Google DeepMind.}  \quad Jason Kuen$^2$ \quad Hao Chen$^1$ \quad Xiaohao Xu$^3$ \\ \quad Jiuxiang Gu$^2$ \quad Yinyi Luo$^1$ \quad Bhiksha Raj$^{1,4}$ \quad Zhe Lin$^{2}$ \quad Marios Savvides$^1$ \\
Carnegie Mellon University$^1$, Adobe Research$^2$, UMich$^3$, MBZUAI$^4$\\
}

\begin{document}
\maketitle

\begin{abstract}
Recent image generation schemes typically capture image distribution in a pre-constructed latent space relying on a frozen image tokenizer. Though the performance of tokenizer plays an essential role to the successful generation, its current evaluation metrics (e.g. rFID) fail to precisely assess the tokenizer and correlate its performance to the generation quality (e.g. gFID). In this paper, we comprehensively analyze the reason for the discrepancy of reconstruction and generation qualities in a discrete latent space, and, from which, we propose a novel plug-and-play tokenizer training scheme to facilitate latent space construction. Specifically, a latent perturbation approach is proposed to simulate sampling noises, i.e., the unexpected tokens sampled, from the generative process. With the latent perturbation, we further propose (1) a novel tokenizer evaluation metric, i.e., pFID, which successfully correlates the tokenizer performance to generation quality and (2) a plug-and-play tokenizer training scheme, which significantly enhances the robustness of tokenizer thus boosting the generation quality and convergence speed. Extensive benchmarking are conducted with 11 advanced discrete image tokenizers with 2 autoregressive generation models to validate our approach. The tokenizer trained with our proposed latent perturbation achieve a notable 1.60 gFID with classifier-free guidance (CFG) and 3.45 gFID without CFG with a $\sim$400M generator. Code: \url{https://github.com/lxa9867/ImageFolder}.
\end{abstract}    
\section{Introduction}
\begin{figure}
    \centering
    \includegraphics[width=\linewidth]{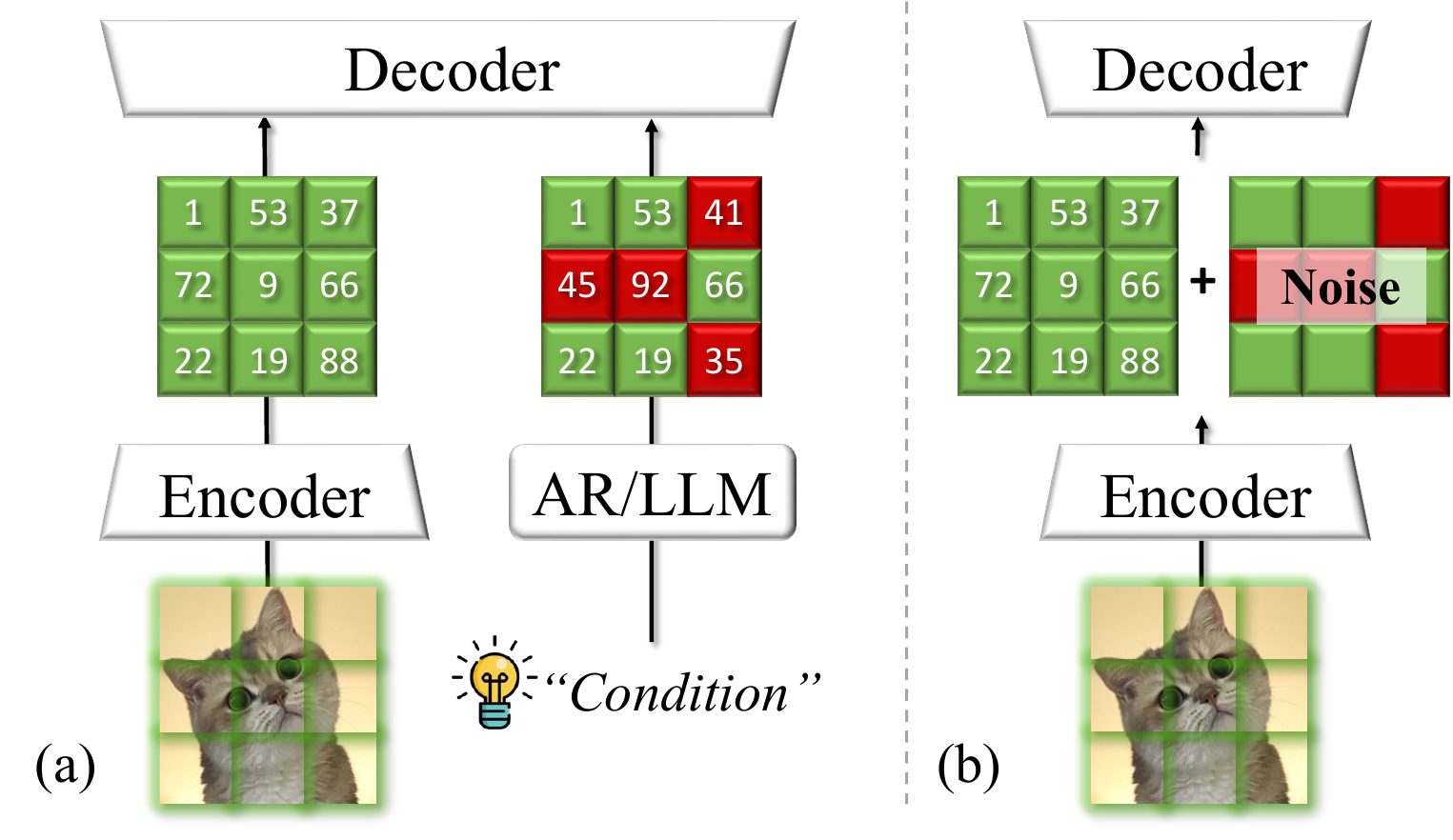}
    \vspace{-0.2in}
    \caption{(a) Traditional image generation scheme: a discrete image tokenizer is fisrt trained with reconstruction target where visual decoder is fed with clean image tokens. After that, an AR/LLM is trained with clean tokens under teacher forcing. However, the during subsequent AR prediction (inference), unexpected tokens can be sampled from the learned distribution and challenge the robustness of frozen visual decoder. (b) RobustTok leverages latent perturbation to enhance the robustness of tokenizer, thus boosting the image generation quality.}
    \label{fig:teaser}
\vspace{-0.2in}
\end{figure}

In recent years, autoregressive (AR) generative modeling \cite{van2016pixel} has gained widespread adoption across various domains, such as text generation \cite{achiam2023gpt}, speech synthesis \cite{chiu2018state,qiu2024efficient}, and image generation \cite{esser2021taming}.
It has emerged as a leading paradigm for the creation of high-quality content \cite{chen2025masked}.
The success of AR models can be largely attributed to a two-stage training process: input tokenization with a tokenizer \cite{lee2022autoregressive,yu2023language,mentzer2023finite,zhu2024scaling,takida2023hq,huang2023towards,zheng2022movqmodulatingquantizedvectors,yu2023magvit,weber2024maskbit,yu2024spae,luo2024open,zhu2024addressing, miwa2025one} and subsequent AR modeling and generation on the discrete latent space of the tokenizer \cite{yu2024randomized,bachmann2025flextok,weber2024maskbit}.

Unlike diffusion models \cite{dhariwal2021diffusionmodelsbeatgans,song2022denoisingdiffusionimplicitmodels}, which operate directly on continuous representations \cite{peebles2023scalablediffusionmodelstransformers,rombach2022highresolutionimagesynthesislatent,vahdat2021scorebasedgenerativemodelinglatent,nichol2021improveddenoisingdiffusionprobabilistic} through a denoising process, AR models typically rely on discrete tokens quantized by the tokenizer.
These discrete tokens facilitate the modeling of latent distribution and subsequent sampling through next-token predictions \cite{van2016pixel, van2016conditional}.
Despite the efficiency in generation \cite{tian2024visualautoregressivemodelingscalable}, AR models suffer significantly from error accumulation \cite{han2024infinity,ren2025beyond}, mainly due to the discrepancies between the training and inference conditions.
Specifically, AR models are usually trained under teacher forcing \cite{sutton1988learning}, where each prediction is based on the preceding ground truth tokens.
In contrast, during inference, predictions rely solely on previously generated tokens, making even minor early errors to propagate and accumulate, and resulting in sampling error, \ie, unexpected tokens being sampled during the AR inference process. 

This issue is further amplified by the typically distinct and misaligned objectives of tokenizer training and AR inference.
Tokenizer training prioritizes reconstruction fidelity where the visual decoder takes \textit{\color{applegreen}{clean}} image tokens for accurate image reconstruction.
Instead, to decoder tokens from a well-trained AR model, the sampling error occurs in the predicted tokens which makes the decoder takes \textit{\color{red}{noisy}} and potentially unseen latent patterns (\cref{fig:teaser} (a)). It necessities the latent space to demonstrate sufficient robustness to handle the potential sampling errors.  
This divergence in objectives also results in a poor correlation between the reconstruction quality metrics \cite{wang2004image}, such as rFID, and the generative performance metrics \cite{heusel2017gans}, such as gFID, an issue repeatedly observed in recent literature \cite{yu2024randomized,weber2024maskbit}.
More recent research also demonstrated that the semantic information contained in latent tokens \cite{li2024imagefolder} and the structure of the latent space \cite{chen2024softvq} can significantly influence the performance of generative models more than reconstruction metrics alone.
Despite these insights, there still lacks a tokenization metric that explicitly captures the quality of the subsequent generation and guides improvements in AR generative modeling.

In this paper, we provide the first comprehensive exploration of how discrete latent space quality affects autoregressive generative modeling. 
Through rigorous analysis, we identify that AR error propagation primarily arises from insufficient robustness in discrete latent spaces.
Motivated by this insight, we propose perturbed FID (pFID), a novel tokenizer evaluation metric designed specifically to measure the discrete latent space robustness under synthesized sampling error.
As shown in \cref{fig:pfid-LlamaGen-B}, pFID effectively correlates and thus predicts downstream generative modeling performance, saving substantial cost to evaluate tokenizers by avoiding generator training.

Building upon these insights, we further introduce a latent perturbation method for tokenizer training (\cref{fig:teaser} (b)) to directly enhance the robustness of latent tokens, thereby significantly mitigating the downstream AR error accumulation and thus improving generation quality.
Specifically, we propose a novel plug-and-play tokenizer training strategy that systematically integrates latent perturbations with an annealing schedule, gradually reducing perturbation intensity to stabilize training and promote robust latent space construction.
Extensive experiments conducted on state-of-the-art autoregressive frameworks, e.g., LlamaGen \cite{sun2024autoregressive} and RAR \cite{yu2024randomized}, across the ImageNet 256x256 generation benchmarks demonstrate the efficacy of our approach. 
Our proposed tokenizer, RobustTok, trained with latent perturbations, substantially outperforms existing methods, achieving notably lower gFID scores with accelerated convergence. 
Furthermore, through detailed ablation studies, we validate the effectiveness of perturbation parameters and confirm that robustness gains directly translate to improved generative performance.

Our contribution can be summarized as follows.
\begin{itemize}
    \item We conduct the first comprehensive analysis identifying insufficient robustness in discrete latent space as a primary factor leading to error accumulation in AR generative modeling.
    \item We propose perturbed FID (pFID), a novel evaluation metric explicitly designed to measure and correlate discrete latent space robustness with downstream generative modeling performance.
    \item We introduce RobustTok, a tokenizer trained using our plug-and-play perturbation approach that achieves superior performance in image generation benchmarks, significantly outperforming existing state-of-the-art methods.
    \item We provide extensive experiments and ablation studies to validate and analyze the effectiveness of latent perturbations in constructing robust discrete latent spaces.
\end{itemize}

\section{Related Works}

\paragraph{Image tokenizers.}
Image tokenization has seen significant advancements across various image-related tasks. Traditionally, autoencoders \cite{hinton2006reducing,vincent2008extracting} have been employed to compress images into latent spaces for downstream applications such as generation and understanding. In generative tasks, VAEs \cite{van2017neural,razavi2019generating} learn to map images to probabilistic distributions; VQGAN \cite{esser2021taming,razavi2019generatingdiversehighfidelityimages} and its subsequent variants \cite{lee2022autoregressive,yu2023language,mentzer2023finite,zhu2024scaling,takida2023hq,huang2023towards,zheng2022movqmodulatingquantizedvectors,yu2023magvit,weber2024maskbit,yu2024spae,luo2024open,zhu2024addressing, miwa2025one} introduce discrete latent spaces to enhance compression and facilitate the application of autoregressive models \cite{vaswani2023attentionneed,dosovitskiy2021imageworth16x16words} to image generation tasks by converting images into sequences of discrete tokens. On the other hand, understanding tasks, such as CLIP \cite{radford2021learning}, DINO \cite{oquab2023dinov2,darcet2023vitneedreg,zhu2024stabilize} and MAE \cite{he2022masked}, rely heavily on LLM \cite{vaswani2023attentionneed,dosovitskiy2021imageworth16x16words} to tokenize images into semantic representations \cite{dong2023peco,ning2301all} where shown its promising performance in classification \cite{dosovitskiy2021imageworth16x16words}, object detection \cite{zhu2010deformable}, segmentation \cite{wang2021maxdeeplabendtoendpanopticsegmentation}, and multi-modal application \cite{yang2024depth}. For a long time, image tokenizer have been divided between methods tailored for generation and those optimized for understanding. Recently, multiple studies have demonstrated the feasibility of leveraging semantic information -- traditionally used for understanding -- in image generation, particularly within tokenization side. \cite{li2024imagefolder,li2024xq,yao2025reconstruction} integrate semantic information into the quantization process and demonstrated its effectiveness in the generative model. On the other hand, recent work \cite{zha2024language, kim2025democratizing,chen2025masked,chen2024softvq} leverages semantic information to mitigate information loss in the high-compression scenarios. 
In this paper, we provide a comprehensive analysis of image tokenizer in a view of perturbation robustness \cite{chen2024slight,li2024r,xuscalable,li2024qdformer,li2023towards,li2023robust}.

\begin{figure*}
    \centering
    \includegraphics[width=0.95\linewidth]{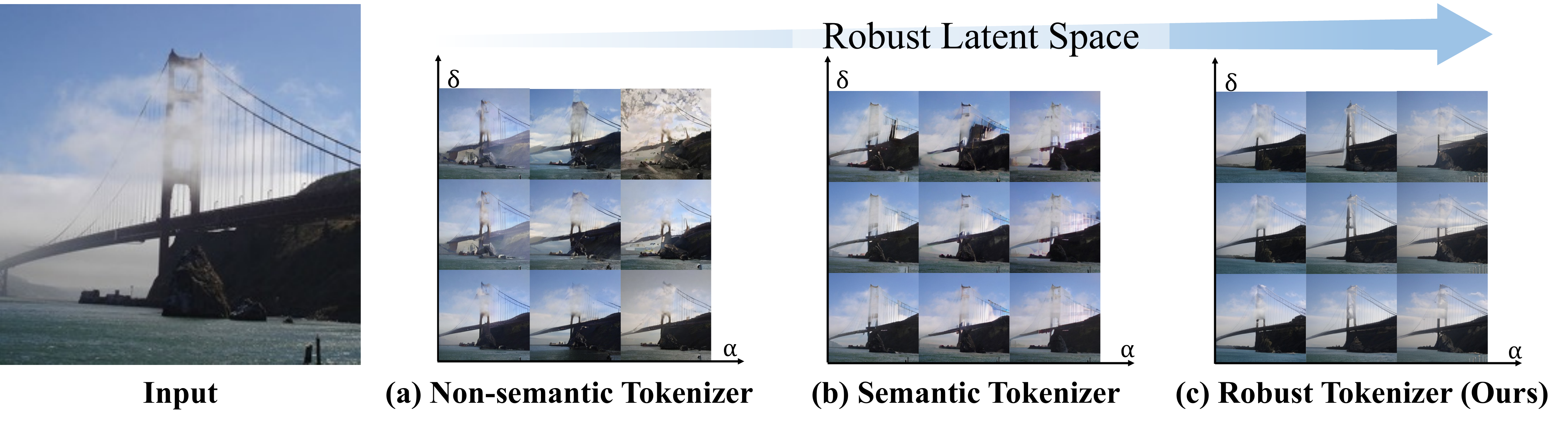}
    \vspace{-0.1in}
    \caption{Visualization of (a) traditional tokenizer, (b) semantic tokenizer, and (c) our RobusTok in reconstruction task with Latent Perturbation. Non-semantic tokenizer leads to distorted reconstructions when perturbations are introduced while our method shows promising robustness to those perturbations.}
    \label{fig:vis-robust}
    \vspace{-0.1in}
\end{figure*}

\paragraph{Autoregressive visual generation.}

Pioneering work on autoregressive visual generation has shown remarkable success in generating high-quality images by modeling the distribution of pixels or latent codes in a sequential manner. The autoregressive model, performed by RNNs \cite{van2016pixel}, CNNs \cite{van2016conditional}, and currently, Transformers \cite{vaswani2023attentionneed}, has demonstrated their strong capacity for capturing long-range dependencies and fine-grained details in image generation. Inspired by exploded development of language model \cite{shi2022divaephotorealisticimagessynthesis, mizrahi20244m} such as GPT \cite{achiam2023gpt}, a series of works leverage tokenizers to convert images or visual information into discrete latent codes, enabling autoregressive or MLM modeling to generate image in raster-scan \cite{esser2021taming} or parallel \cite{pang2024next,chang2022maskgitmaskedgenerativeimage,wang2024parallelized} order. Recently, autoregressive models continued to show their scalability power in larger datasets and multimodal tasks \cite{he2024mars}; models like LlamaGen \cite{sun2024autoregressive} and AiM \cite{li2024scalable} adapting current advanced LLM architectures for image generation. New directions such as VAR \cite{tian2024visualautoregressivemodelingscalable,li2024controlvar,han2024infinity,ren2024flowar,qiu2024efficient} and RAR \cite{yu2024randomized,pang2024randar} focus on fusing global information into the training of autoregressive model and successfully achieve a promising result. MAR \cite{li2024autoregressiveimagegenerationvector}, Fluid \cite{fan2024fluid}, and GIVIT \cite{tschannen2024givt} have shown the potential for continuous image generation through autoregressive modeling. Based on the development of such, various technique are continue to unify the language language model for generation and understanding \cite{wu2024liquid,tong2024metamorph}.

\section{Sampling Error Synthesis}

Motivated by the AR improvement over semantic tokenization, we analyze the significance of sampling error during AR inference and propose a latent perturbation method for tokenizer training in this section.

\subsection{Preliminaries}

\paragraph{Vector Quantization (VQ).}
Most AR models are based on discrete tokenizers with a quantized latent space. 
The tokenizer usually consists of an encoder, a quantizer, and a decoder.
Although many quantization techniques for the quantizer were previously proposed \cite{mentzer2023finite,yu2023language,zhao2024image}, we focus on the VQ tokenizer \cite{esser2021taming} for its simplicity and natural compatibility with AR models in this paper.

Given an RGB image $I$, the encoder $\mathcal{E}$ first extracts a set of latent representations $Z \in \mathbb{R}^{H \times W \times C}$, where $H \times W$ denotes the spatial resolution of the latent tokens.
VQ \cite{esser2021taming} aims to quantize continuous features into a set of discrete features $Z'$ with a minimum reconstruction error of the original data, ensuring that the quantized representation remains as close as possible to the original continuous ones. 
Specifically, it maps each continuous feature vector $\mathbf{z} \in \mathbb{R}^C$ to a closest quantized codeword $\mathbf{e} \in \mathbb{R}^C$ from a learnable codebook $\mathcal{C}=\{\mathbf{e}_k\}_{k=1}^K$ with in total $K$ codewords as:
\begin{equation}
    \mathbf{z}^\prime=\arg\min_{\mathbf{e}_k\in\mathcal{C}}\|\mathbf{z}-\mathbf{e}_k\|_2^2.
\end{equation}
The decoder $\mathcal{D}$ then reconstructs the original input signals by taking the quantized representation $Z'$ as input.

\paragraph{Autoregressive Generation with VQ.} 
Given a sequence of quantized tokens $Z' = \{\mathbf{z}'_1, \cdots, \mathbf{z}'_{T}\}$ of length $T = H \times W$, AR models capture the entire distribution as:
\begin{equation}
    p(\mathbf{z}'_1,\cdots,\mathbf{z}'_T;\theta)=\prod_{t=1}^T p(\mathbf{z}'_t|\mathbf{z}'_{<t};\theta),
\end{equation}
where $\theta$ represents the parameters of a deep neural network. 
To learn the network, AR models are trained to predict the tokens at timestep/position $t$, given all the preceding ground truth tokens, known as teacher forcing \cite{sutton1988learning}.
However, this mechanism introduces a discrepancy in the inference stage, where, during AR inference, the predicted tokens are conditioned on the preceding predictions instead of the ground truth ones. 
This discrepancy can introduce and then accumulate errors during inference, resulting in generations of degraded quality \cite{bengio2015scheduled,lamb2016professor}.

\begin{figure*}
    \centering
    \includegraphics[width=\linewidth]{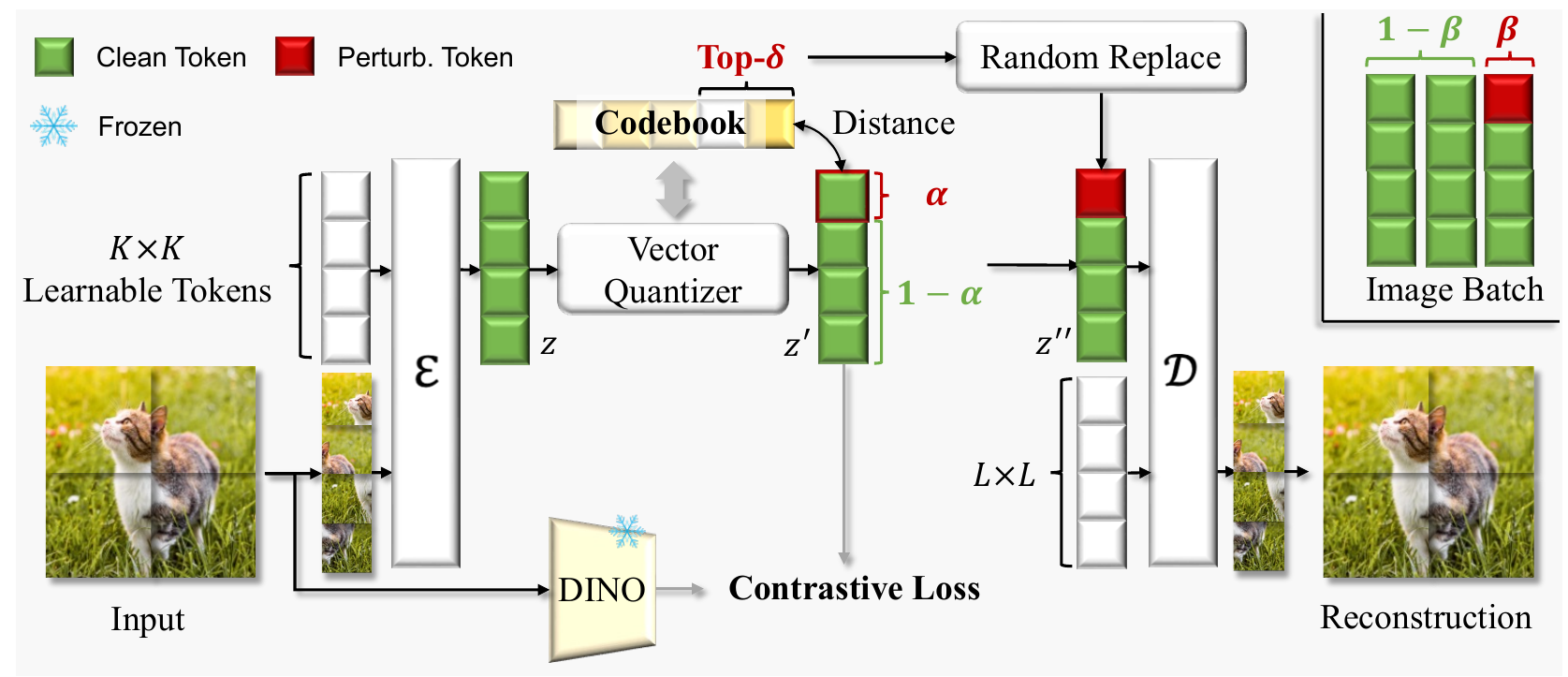}
    \caption{RobustTok overview. We adopt vision transformer as our encoder $\mathcal{E}$ and decoder $\mathcal{D}$. $\beta$ of data in one batch will process our Latent Perturbation, which will be randomly replaced by top-$\delta$ neighbor from codebook with probability $\alpha$. A frozen DINO encoder is utilized to supervise our latent space. }
    \label{fig:pipeline}
\vspace{-0.4cm}
\end{figure*}

\subsection{Latent Perturbation}

Knowing that AR models are subjected to sampling error accumulation due to the discrepancy between training and inference, we show in the following that such sampling error of AR models can be captured within the tokenizer alone with a novel reconstruction metric, and can be mitigated by involving perturbation into tokenizer training in a plug-and-play manner.
More specifically, we can simulate the sampling error, \ie, unexpected tokens during sampling of AR models, with perturbed latent tokens and to enhance the robustness of the tokenizer, as shown in \cref{fig:vis-robust}.

\paragraph{Perturbation rate.}
An important metric to monitor the AR modeling process is the accuracy of predicted tokens. Likewise, we define a perturbation rate $\alpha$ to control the proportion of perturbed token within an image. Given the quantized feature $Z^\prime\in\mathbb{R}^{H\times W \times C}$, we define $\alpha$ as:
\begin{equation}
    \alpha = \frac{P}{H\times W},
\end{equation}
where $P$ denotes the perturbed token number.
To simulate the sampling error, we can randomly perturb the quantized tokens from the tokenizer encoder.

\paragraph{Perturbation proportion.}
Within a batch of images, we apply the perturbation in a proportion $\beta$ of images and keep the remaining images unchanged. With $N_c$ clean images and $N_p$ perturbed images, the perturbation proportion is calculated as:
\begin{equation}
    \beta = \frac{N_c}{N_c+N_p}.
\end{equation}

\paragraph{Perturbation strength.}
We define a perturbation strength $\delta$ to quantify the perturbation level. Specifically, given a discrete token $\mathbf{z}=\mathbf{e}_k$ with a codebook $\mathcal{C}$, we calculate the set of top-$\delta$ nearest neighbors:
\begin{equation}
   \mathcal{S}_\delta = \underset{\mathcal{S}_\delta \subset \mathcal{C}, |\mathcal{S}_\delta|=\delta}{\arg\min} \sum_{\mathbf{e}_n \in \mathcal{S}_\delta} \|\mathbf{e}_n - \mathbf{e}_k\|_2^2,
\end{equation}
where $|\cdot|$ denotes the counting operation. 
We randomly replace the original token $\mathbf{e}_k$ with a $\mathbf{e}_\delta\in\mathcal{S}_\delta$ to perturb the latent, thereby modifying the latent representation to simulate sampling in AR with the top-k nucleus strategy.

\subsection{Robustness Indicator: Perturbed FID}

\begin{table}[t]
    \centering
    \begin{tabular}{p{2.2cm}<{\centering}|p{2.25cm}<{\centering}|p{2.2cm}<{\centering}}
     Ideal Scenario & Train Tokenizer & Eval AR/LLM \\
     \hline
     $\mathcal{D}(\mathbf{z})=I$ & $\mathcal{D}(\mathbf{z})=\hat{I}$ & $\mathcal{D}(\mathbf{z}+\Delta)=\hat{I}^\prime$ \\ 
    \end{tabular}
    \caption{Decoder analysis. $I$: input image. $\hat{I}$: predicted image. $\hat{I}^\prime$: predicted image from noisy latent. $z$: latent feature. $\Delta$: sampling error. $\mathcal{D}$: decoder.}
    \label{tab:analysis}
\vspace{-0.15in}
\end{table}

\begin{table*}
\begin{center}
\renewcommand{\arraystretch}{1.2} 
\setlength{\tabcolsep}{9pt} 
\scalebox{0.9}{
\begin{tabular}{c|
p{3cm}<{\centering}
p{3.cm}<{\centering}|
p{1.5cm}<{\centering}
p{1.5cm}<{\centering}|
p{1.5cm}<{\centering}
p{2cm}<{\centering}}
\toprule
\hline
\multirow{2}{*}{Codebook Size} & \multirow{2}{*}{Method} & \multirow{2}{*}{Tokenizer Type} & \multicolumn{2}{c}{Tokenizer} & \multicolumn{2}{c}{Generator} \\
    \cline{4-7}
    & & & rFID$\downarrow$ & pFID$\downarrow$ & gFID$\downarrow$ & gFID$\downarrow$ (CFG) \\

\midrule
\multirow{4}*{16384}& VQGAN-16384 \cite{esser2021taming} & Non-semantic & 4.50 & 18.18 & 37.39 & 14.80 \\
~ & LlamaGen \cite{sun2024autoregressive} & Non-semantic & 2.19 & 13.12 & 26.34 & 8.61 \\
& IBQ-16384 \cite{shi2024taming} & Non-semantic & 1.41 & 16.35 & 30.19 &  11.01 \\
~ & VQGAN-LC \cite{zhu2024scaling} & Semantic$^*$ & 3.27 & 16.78 & 31.35 & 11.80 \\
\midrule
\multirow{3}*{8192} & IBQ-8192 \cite{shi2024taming} & Non-semantic & 1.87 & 19.62& 30.91 & 10.85 \\
~ & TiTok \cite{yu2024imageworth32tokens} & Semantic &  1.03 & 3.55 & 25.66 & 8.84 \\
~ & XQGAN-8192 \cite{li2024xq} & Semantic & \textbf{0.81} & 7.91 & 25.43 & 10.18 \\
\midrule
\multirow{2}*{4096} & XQGAN-4096 \cite{li2024xq} & Semantic & 0.91 & 6.98 & 13.58 & 6.91 \\
 & \cellcolor{blue!8}RobustTok (Ours) & \cellcolor{blue!8}Semantic + Robust & \cellcolor{blue!8}1.02 & \cellcolor{blue!8}\textbf{2.28} & \cellcolor{blue!8}\textbf{9.47} & \cellcolor{blue!8}\textbf{5.67} \\
\midrule
\multirow{2}*{1024} &  MaskGIT \cite{chang2022maskgitmaskedgenerativeimage} & Non-Semantic & 2.28 & 4.20 & 18.02 & 5.85\\
& IBQ-1024 \cite{shi2024taming} & Non-Semantic & 2.24 & 6.37 & 35.33 & 11.01\\
\hline
\bottomrule
\end{tabular}
}
\vspace{-0.2cm}
\caption{Benchmark of tokenizers with the same LlamaGen-B generator. For fair comparison, the gFID with classifier-free guidance utilizes the same classifier value and schedule. All the tokenizers share the same $C\times 16\times 16$ latent shape. We discuss the reason of choosing codebook size 4096 to train RobustTok in the ablation. More benchmarking results with larger generators are available in the appendix. $^*$ denotes semantics captured with linear projection. All metrics, \ie, rFID, pFID and gFID, are the smaller the better.}
\vspace{-0.1in}
\label{tab:token_setting}

\end{center}    
\end{table*}

\paragraph{Analysis - Lipschitz smoothness.}
We analyze the discrepancy between tokenizer training and inference schemes. As shown in \cref{tab:analysis}, we demonstrate the input/output formulation of visual decoder upon the latent representations $\mathbf{z}$. Ideally, the decoder $\mathcal{D}$ should take a clear latent $\mathbf{z}$ and reconstruct the input image $I$ that aligns with the current tokenizer training target. However, during the inference stage with a well-trained AR/LLM, sampling error $\Delta$ always happens. This will change the usage of the decoder differently from its training target, which significantly challenges the robustness of the visual decoder during inference as we expect $\mathcal{D}(\mathbf{z}+\Delta)$ can still reconstruct the input $I$. The decoder's robustness can be measured by the Lipschtz smoothness $Lip=\frac{\hat{I}^\prime-\hat{I}}{\Delta}\approx\frac{\hat{I}^\prime-I}{\Delta}$. However, in a discrete latent space, the potential choice of $\Delta$ is constrained and the discrepancy between input $I$ and reconstructed images $\hat{I}^\prime$ can be better reflected by the Fréchet Inception Distance (FID). In this way, we introduce perturbed FID (pFID) as a new metric to reflect the robustness and reconstruction quality of image tokenizers.

\paragraph{Perturbed FID.}
With the latent perturbation parameters: $\alpha$, $\beta$ and $\delta$, we propose a novel reconstruction metric, termed as Perturbated FID (pFID).
Compared to reconstruction FID (rFID) that merely captures the reconstruction quality of the tokenizer, pFID can reflect the robustness and the latent space from a tokenizer, and correlates with the sampling error and thus the performance of AR models.

To calculate the pFID, we apply perturbation among all images, \ie, $\beta=1$ for all the settings. In addition, to simulate different noisy level during inference, we define a set of perturbation rates $\alpha\in\{0.9, 0.8, 0.7, 0.6, 0.5\}$ and a set of perturbation strength $\delta\in\{200, 280, 360\}$. Combining both sets, we have a total of 15 combinations of perturbation settings. We generate images with all settings and calculate the FID between input images. The averaged value serves as pFID. Specifically, the perturbation strength is linearly scaled to adopt different codebook sizes. 

We present a comparison of rFID and pFID in \cref{fig:pfid-LlamaGen-B}, and provide more analysis on the results in \cref{sec:exp-results}.
To summarize, our pFID is more correlated with the tokenizer's downstream generation performance compared with rFID.

\begin{table*}[t]
\centering
\small
\renewcommand{\arraystretch}{1.0} 
\setlength{\tabcolsep}{9pt} 
\begin{tabular}{l| l|lc|cccc|cccc}
\toprule
\hline
\multirow{2}{*}{Type} & \multirow{2}{*}{Method} & \multicolumn{2}{c|}{Tokenizer} & \multicolumn{7}{c}{Generator} \\
\cline{3-11}
 ~ & ~ & rFID$\downarrow$ & pFID$\downarrow$ & gFID$\downarrow$ & IS$\uparrow$ & Pre$\uparrow$ & Rec$\uparrow$ & \#Para & Leng. & Step \\
\hline
\multirow{4}{*}{Diff.} 
 & ADM \cite{dhariwal2021diffusionmodelsbeatgans} & - & - & 10.94 & 101.0 & 0.69 & 0.63 & 554M & - & 1000 \\
 & LDM-4 \cite{rombach2022highresolutionimagesynthesislatent} & - & - & 3.60 & 247.7  & - & - & 400M & - & 250 \\
 & DiT-L/2 \cite{peebles2023scalablediffusionmodelstransformers} & 0.90 & - & 5.02 & 167.2 & 0.75 & 0.57 & 458M & - & 250 \\
 & MAR-B \cite{li2024autoregressiveimagegenerationvector} & 1.22 & - & 2.31 & 281.7 & 0.82 & 0.57 & 208M & - & 64 \\
\midrule
\multirow{5}{*}{NAR} 
 & MaskGIT \cite{chang2022maskgitmaskedgenerativeimage} & 2.28 & 5.03 & 6.18 & 182.1 & 0.80 & 0.51 & 227M & 256 & 8 \\
 & RCG (cond.) \cite{li2024returnunconditionalgenerationselfsupervised} & - & - & 3.49 & 215.5 & - & - & 502M & 256 & 250 \\
 & TiTok-S-128\cite{yu2024imageworth32tokens} & 1.52 & - & 1.94 & - & - & - & 177M & 128 & 64 \\
 & MAGVIT-v2 \cite{yu2023language} & 0.90 & - & 1.78 & 319.4 & - & - & 307M & 256 & 64 \\
 & MaskBit \cite{weber2024maskbit} & 1.51 & - & 1.65& 341.8 & - & - & 305M& 256 & 64\\
\midrule
\multirow{8}{*}{AR} 
 & VQGAN \cite{esser2021taming} & 7.94 & - & 18.65 & 80.4 & 0.78 & 0.26 & 227M & 256 & 256 \\
 & RQ-Transformer \cite{lee2022autoregressiveimagegenerationusing} & 1.83 & - & 15.72 & 86.8 & - & - & 480M & 1024 & 64 \\
 & LlamaGen-L \cite{sun2024autoregressive} & 2.19 & 13.12 & 3.80 & 248.3 & 0.83 & 0.52 & 343M & 256 & 256 \\
 & VAR \cite{tian2024visualautoregressivemodelingscalable} & 0.90 & 17.46 & 3.30 & 274.4 & 0.84 & 0.51 & 310M & 680 & 10 \\
 & ImageFolder \cite{tian2024visualautoregressivemodelingscalable} & 0.80 & 7.23 & 2.60 & 295.0 & 0.75 & 0.63 & 362M & 286 & 10 \\
 & RAR-B \cite{yu2024randomized} & \multirow{2}{*}{2.28} &  \multirow{2}{*}{5.03} & 1.95 & 290.5 & 0.82 & 0.58 & 261M & 256 & 256 \\
& RAR-L \cite{yu2024randomized} &  & & 1.70 & 299.5 & 0.81 & 0.60 & 461M & 256 & 256 \\
\rowcolor{blue!8} \cellcolor{white} & RobustTok-RAR-B (Ours) &   &   & 1.83 & 298.3 & 0.80 & 0.63 & 261M & 256 & 256 \\
\rowcolor{blue!8} \cellcolor{white} & RobustTok-RAR-L (Ours) & \multirow{-2}{*}{1.02} & \multirow{-2}{*}{2.28} & 1.60 & 305.8 & 0.78 & 0.65 & 461M & 256 & 256 \\
\hline
\bottomrule
\end{tabular}
\caption{System-level performance comparison on class-conditional ImageNet 256x256. $\uparrow$ and $\downarrow$ indicate that higher or lower values are better, respectively.}
\vspace{-0.1in}
\label{tab:main-result}
\end{table*}

\subsection{RobustTok}

Inspired by the proposed pFID metric, which shows the robustness of the discrete latent space is important to capture the sampling error of AR models, we demonstrate that we can further involve such perturbation into tokenizer training to proactively learn a more robust latent space.

\paragraph{Architecture.}
Following prior works \cite{li2024imagefolder, li2024xq, yu2024imageworth32tokens}, RobustTok leverages Vision Transformer (ViT) \cite{dosovitskiy2020image} as visual encoder and visual decoder. As shown in \cref{fig:pipeline}, we initialize a set of learnable tokens and use these tokens as the representation for image reconstruction and subsequent generation. Specifically, the input image is first patchified to $L\times L$ tokens, where $L$ represents the patch size, and concatenated with learnable tokens to serve as the input of the encoder. We apply vector quantization on the continuous token $z$ obtained from the encoder $\mathcal{E}$. After that a latent perturbation approach is applied to guide the latent space construction. Finally, the ViT decoder takes perturbed tokens $z^{\prime\prime}$ and a new set of learnable tokens to reconstruct the image. Specifically, we incorporate a pretrained DINOv2 model \cite{oquab2023dinov2} to inject semantics, ensuring that the learned tokens retain meaningful visual semantics and structural coherence.

\paragraph{Plug-and-play perturbation.}
During tokenizer training, we apply latent perturbation to enhance its robustness. We apply perturbation after semantic regularization \cite{li2024imagefolder} to preserve clear semantics in the discrete tokens to maximize the reconstruction capability. Within a batch of image, we randomly choose $\beta$ of them to add perturbation. To apply perturbation to each selected image, we randomly choose $\alpha\times H\times W$ tokens and then calculate the top-$\delta$ nearest neighbors to those tokens within the learned codebook. 
The final perturbation is applied by randomly replacing the original token with its top-$\delta$ nearest neighbor.  


\section{Experiment}

\subsection{Experimental Setting}
We experiment on 256$\times$256 ImageNet \cite{deng2009imagenet} benchmark for both reconstruction and generation. 
As summarized in \cref{tab:token_setting}, we first evaluate 11 popular tokenizers across 4 codebook sizes. This almost includes all open-sourced discrete tokenizers. 
The usage of all off-the-shelf tokenizers follows their official implementation and pre-trained weights. 
We pre-tokenize images with pre-trained tokenizers and benchmark their generation performance using LlamaGen-Base/Large generators with default settings \cite{sun2024autoregressive}. 
For RobustTok, we additionally leverage RAR \cite{yu2024randomized} as an additional generator to validate its wide applicability.

\paragraph{Evaluation metric.} We employ Fréchet Inception Distance (FID) \cite{heusel2017gans}, Inception Score (IS) \cite{salimans2016improved}, Precision, and Recall as metrics to assess generation quality. 
We report the results of both using classifier-free guidance (CFG) \cite{ho2022classifier} and without using CFG.
For tokenizer, we utilize rFID and pFID to evaluate tokenizer reconstruction quality and robustness. 

\begin{figure*}
    \centering
    \begin{subfigure}{0.48\linewidth}
        \centering
        \includegraphics[width=\linewidth]{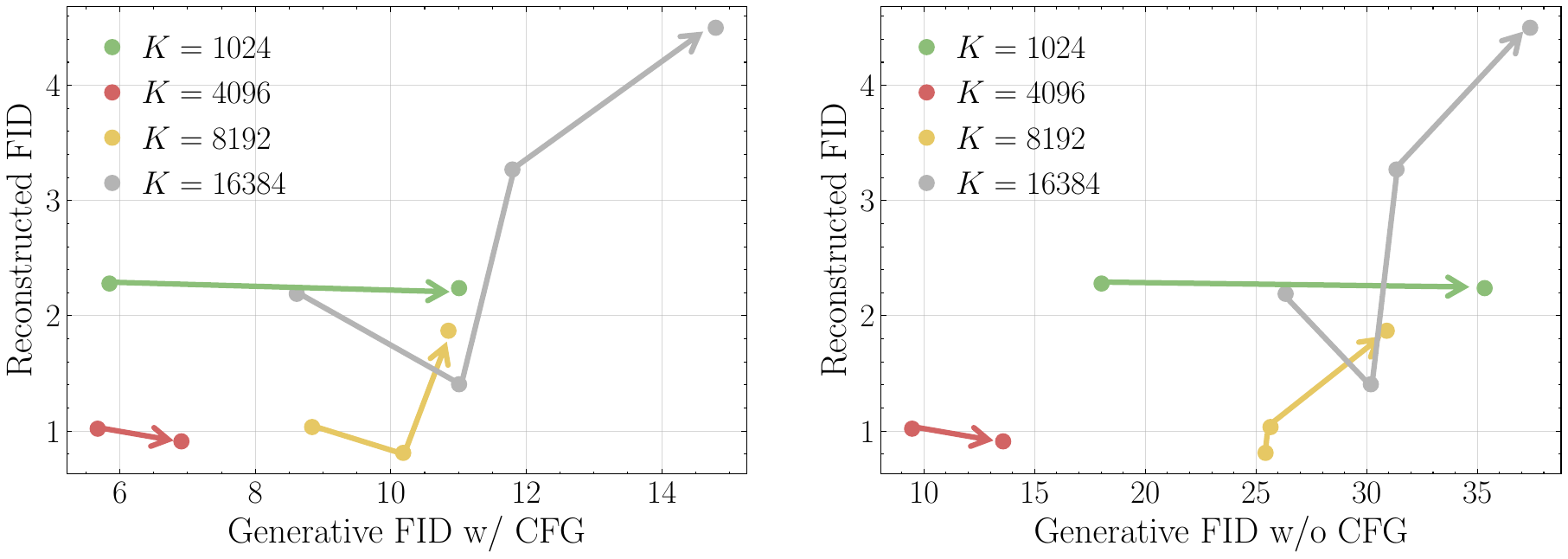}
        \caption{\textbf{rFID} v.s gFID with and without CFG.}
        \label{fig:pfid-LlamaGen-B-recon}
    \end{subfigure}
    \hfill
    \begin{subfigure}{0.48\linewidth}
        \centering
        \includegraphics[width=\linewidth]{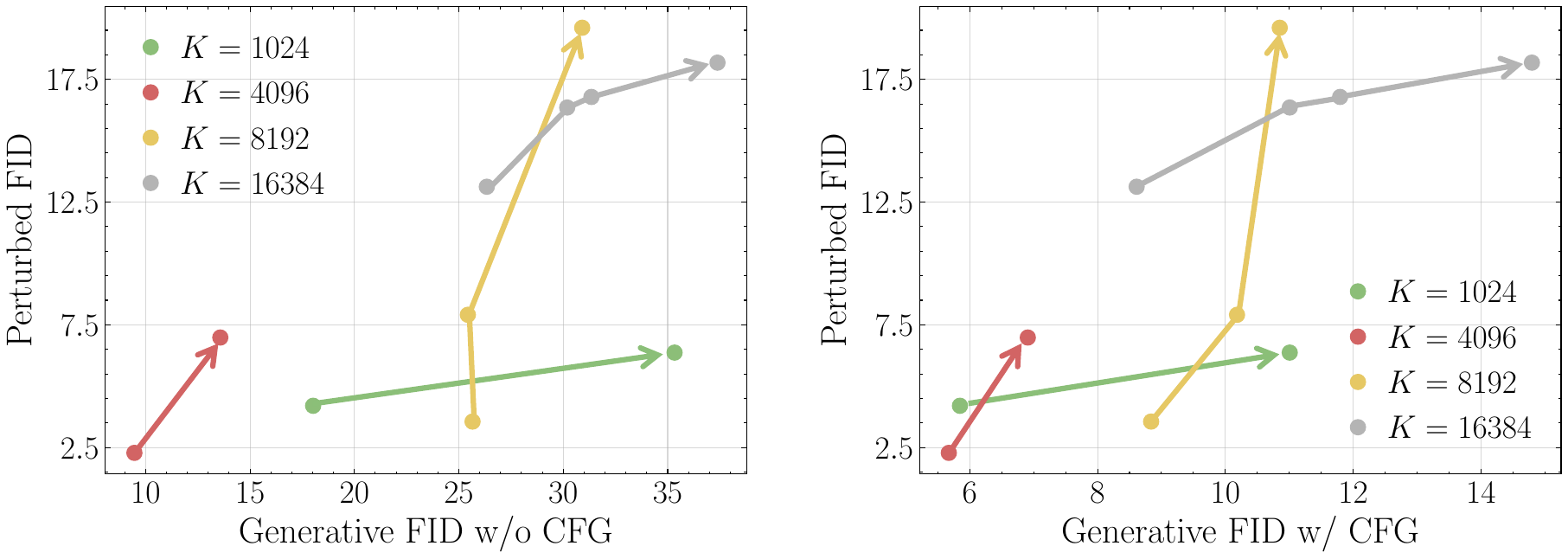}
        \caption{\textbf{pFID} vs. gFID with and without CFG.}
        \label{fig:pfid-LlamaGen-B-pert}
    \end{subfigure}
    \caption{Comparison of rFID-gFID and pFID-gFID curves of differnt tokenizers under LlamaGen-B training setting. $K$ denotes codebook size. Each point represents a method in \cref{tab:token_setting}.}
    \vspace{-0.1in}
    \label{fig:pfid-LlamaGen-B}
\end{figure*}

\paragraph{Implementation details.}  
RobustTok follows the XQGAN \cite{li2024xq} training setting but replaces product quantization with vanilla vector quantization \cite{li2024imagefolder}. 
We retain semantic regularization to stabilize the latent space. During tokenizer training, we randomly select $\beta = 0.1$ of the total data to add perturbation. For these selected samples, we set $\alpha = 1.0$ and $\delta = 100$, and gradually anneal to half over the training. For the AR generator, we strictly follow the training recipes of LlamaGen \cite{sun2024autoregressive} and RAR \cite{yu2024randomized} except for changing the tokenizer to RobustTok. 

\subsection{Result Analysis}
\label{sec:exp-results}

\paragraph{Generic take-home observations.}
Before we go through and validate the core focus of this paper, we aim to conclude some generic observations from the benchmarking. The observations are summarized from the benchmarking results of LlamaGen-Base/Large.
\begin{itemize}
    \item \textbf{Codebook size}: With similar reconstruction capability, the smaller the codebook size, the better the generation quality. We consider this property primarily results from the simple latent space are easier to capture during the AR modeling.
    \item \textbf{Semantics}: Semantic tokenizer typically demonstrates better capability for both reconstruction and generation. Semantic guidance provides a structural and clustering latent for better compression capability for reconstruction and robustness property for generation accordingly.
    \item \textbf{Reconstruction}: Reconstruction capability measured by traditional rFID does not align with the generation capability. This should be potentially resulted from the discrepancy between tokenizer training and inference, \ie, the latent space lacks robustness.
\end{itemize}

\paragraph{Effectiveness of pFID.}
To better compare the correlation among metrics, we visualize the rFID-gFID and pFID-gFID curves as shown in \cref{fig:pfid-LlamaGen-B} (more results with the LlamaGen-Large generator are available in the Appendix). (a) When comparing rFID and gFID, we observe that there is no clear correlation between them, regardless of whether classifier-free guidance is used in generation or not. (b) Differently, pFID and gFID demonstrate a strong correlation within each codebook size $K$. We separately compare results within each $K$ primarily because we add different perturbation strength $\delta$ according to $K$. With the new pFID, we can better access the tokenizer's performance without the time-consuming and resource-intensive training of subsequent generators.

\begin{figure}
    \centering
    \includegraphics[width=0.95\linewidth]{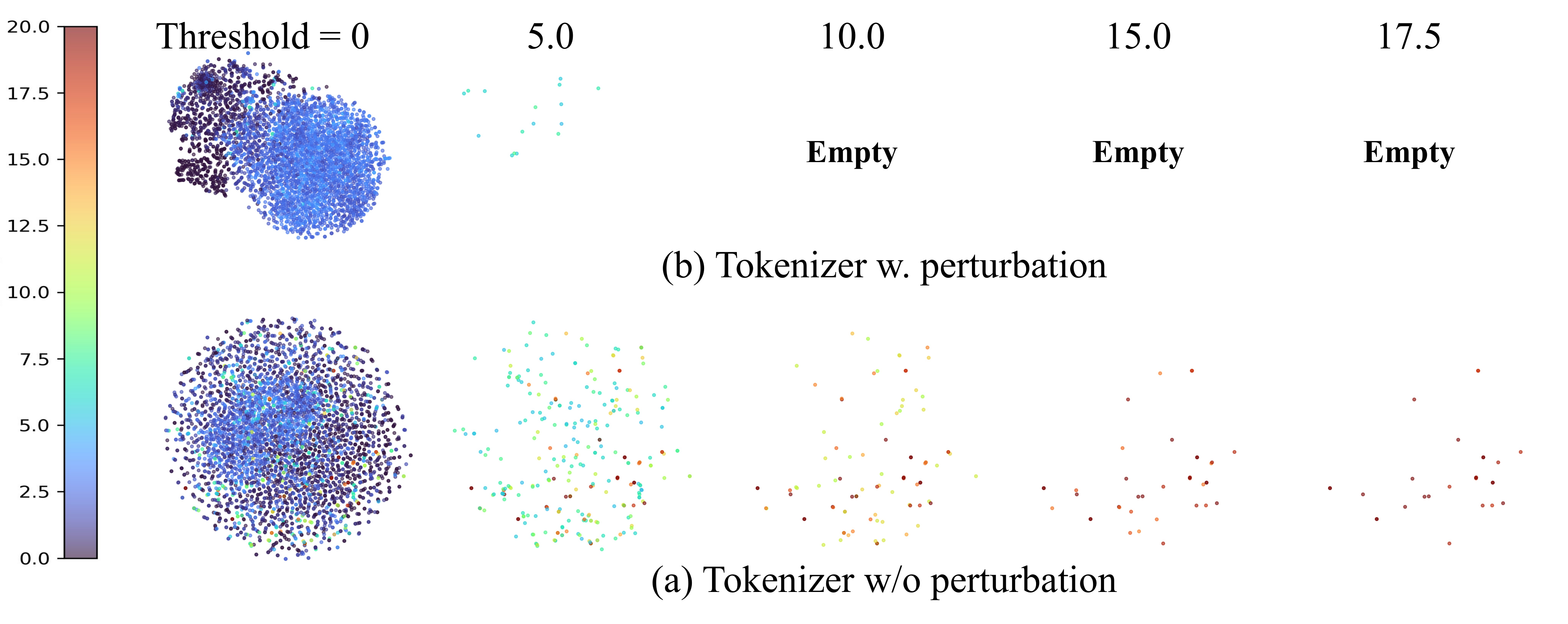}
    \caption{T-SNE visualization of latent space of tokenizer trained with and without latent perturbation. Colors and thresholds represent the frequency of tokens being used during inference without perturbation.}
    \label{fig:tsne-intuition-base}
    \vspace{-0.1in}
\end{figure}

\definecolor{NoneColor}{HTML}{f57c6e}
\definecolor{HalfColor}{HTML}{84c3b7}
\definecolor{ZeroColor}{HTML}{b8aeeb}

\paragraph{Systematic comparison.}
As shown in \cref{tab:main-result}, we compare our RobustTok with various state-of-the-art methods on the ImageNet 256x256 \cite{deng2009imagenet} benchmark. Notably, our proposed RobustTok leads to a significant improvement over previous methods. Specifically, 0.12 and 0.10 gFID gains are achieved by utilizing RobustTok on top of RAR generator. And finally, with a 461M model, our approach achieves a 1.60 gFID. 

\paragraph{Robust latent space.}
As shown in \cref{fig:tsne-intuition-base}, we compare the latent space (\ie, codebook) with and without latent perturbation. We colorize the latent tokens with their frequency of use during inference. When truncating tokens at different usage count thresholds, we observe the space constructed with latent perturbation contains many reusable tokens, which acted as key tokens that can be easily modeled, while the remaining tokens serve as supportive tokens providing finer detailed information. In contrast, the latent space without latent perturbation distributes usage more uniformly across tokens. 

\begin{figure}
    \centering
    \includegraphics[width=1.0\linewidth]{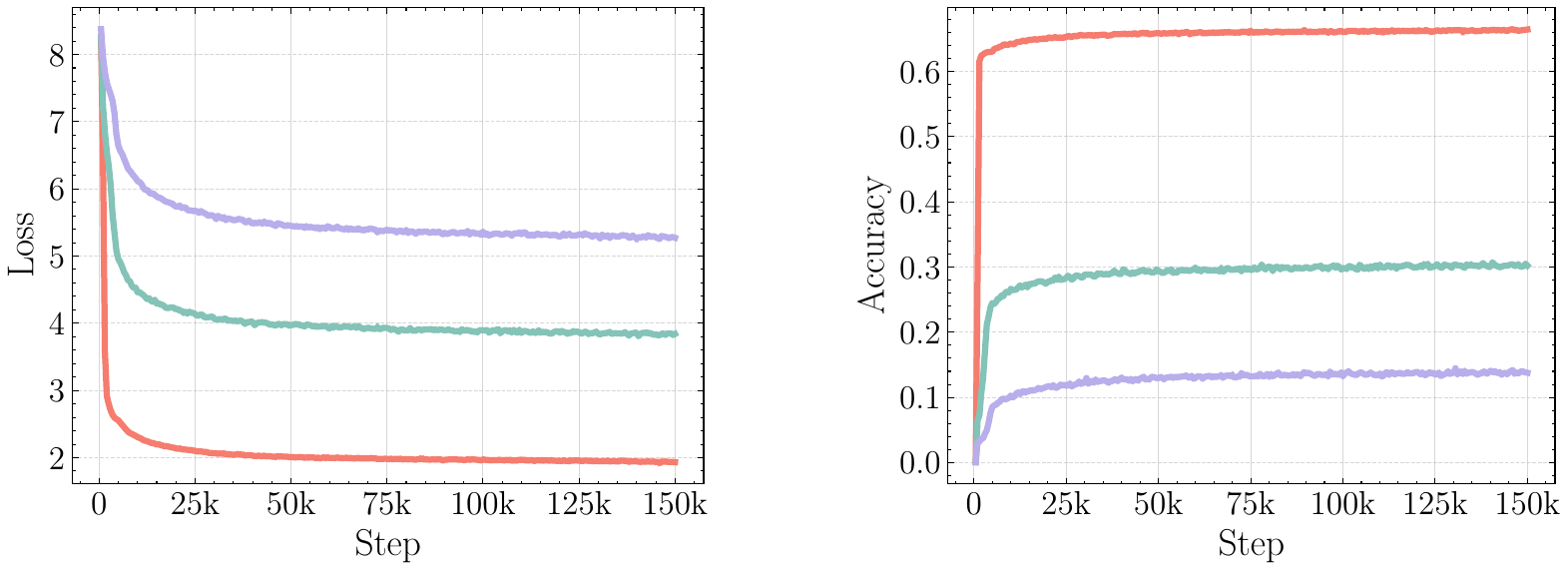}
    \caption{RAR training curve for \textcolor{NoneColor}{None}, \textcolor{HalfColor}{Half}, and \textcolor{ZeroColor}{Zero} annealing strategies. The tokenizer without annealing exhibits strong convergence but compromises diversity, while annealing to zero offers limited improvement over the baseline.}
    \vspace{0.1in}
    \label{fig:loss-curve}
\end{figure}

\begin{figure*}[t]
    \centering
    \includegraphics[width=\textwidth]{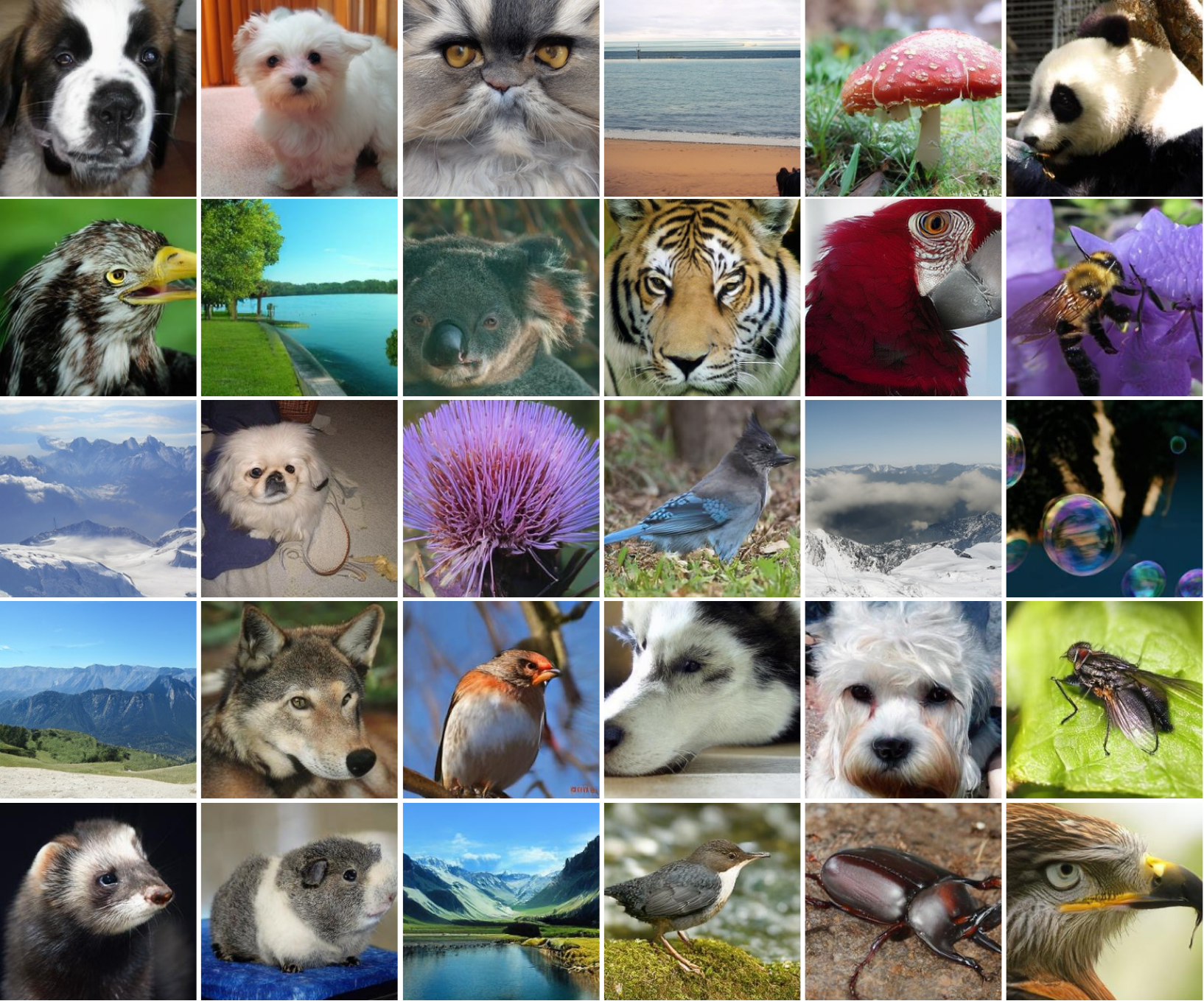} 
    \vspace{-0.25in}
    \caption{Visualization of 256 × 256 image generation within ImageNet class.}
    \vspace{-0.15in}
    \label{fig:visualization}
\end{figure*}

\begin{table}
\centering
\setlength{\tabcolsep}{6pt} 
\resizebox{0.95\linewidth}{!}{%
\begin{tabular}{l|cc|cc}
\toprule

\multirow{2}{*}{Method} & \multirow{2}{*}{rFID$\downarrow$} & \multirow{2}{*}{pFID$\downarrow$} & \multicolumn{2}{c}{gFID$\downarrow$} \\
& & & w./o. CFG & CFG \\
\midrule
\multicolumn{5}{l}{\textit{LlamaGen-L}\vspace{0.02in}} \\
\pz\pz XQGAN \cite{li2024xq} & 0.81 & 7.91 & 14.64 & 4.48 \\
\pz\pz\pz + $K$ = 4096 & 0.91 & 6.98 & 7.91 & 4.13\\ 
\pz\pz\pz  + L.P. $\beta = 0.5$ & 3.97 & 4.52 & 9.31 & 5.40 \\
\pz\pz\pz  + L.P. $\beta = 0.1$ & 1.58 & 3.61 & 4.60 & 3.93 \\ 
\midrule
\multicolumn{5}{l}{\textit{RAR-B}\vspace{0.02in}} \\
\pz\pz\pz  + L.P. $\mapsto$ 0 & 0.97 & 4.89 & 5.32 & 1.97 \\ 
\pz\pz\pz  + L.P. $\mapsto$ 0.5 & 1.02 & 2.28 & 4.62 & 1.85 \\ 
\bottomrule
\end{tabular}
}
\vspace{-0.05in}
\caption{Ablation of RobustTok. L.P. stands for our proposed Latent Perturbation. $\mapsto0/0.5$ denotes annealing the perturbation to none and half respectively. gFID with classifier-free guidance (CFG) uses the constant schedule for LlamaGen and the linear schedule for RAR.}
\vspace{-0.1in}
\label{tab:ablation}
\end{table}


\paragraph{Perturbation selection \& annealing strategy.}

As shown in \cref{tab:ablation}, we conduct an ablation to determine the optimal selection of perturbation hyperparameters. Our results indicate that using a large perturbation parameter, \eg, $\beta = 0.5$, degrades the model's reconstruction capability and adversely affects generative performance. Furthermore, training without annealing strategy leads to mode collapse and loss of generation diversity, whereas annealing to zero results in an overly deterministic tokenizer, diminishing the flexibility observed in \cref{fig:tsne-intuition-base}. We find that annealing to half strikes a balance between robustness and adaptability, preserving essential latent properties while improving the quality of generated outputs. We show the loss curves and accuracy of predicted tokens during training in \cref{fig:loss-curve}.

\paragraph{Qualitative results.}
We demonstrate images generated by our approach as shown in \cref{fig:visualization}.

\section{Conclusion}

\textbf{Limitation.} 
Though we focus on a discrete latent space in this paper, the discussed problem also exists in continuous tokenizers with diffusion models. However, the latent perturbation (\eg, non-scheduled noise) in continuous tokenizer is more challenging to determine as the perturbation is not constrained by a codebook. We leave this as future work.

In this paper, we introduce RobustTok, a novel tokenizer training scheme designed to enhance the robustness of discrete latent spaces in autoregressive image generation. Through our proposed latent perturbation approach, we successfully address the issue of error accumulation that arises from discrepancies between training and inference conditions. Furthermore, we introduced Perturbed FID (pFID), a new metric that effectively correlates tokenizer robustness with downstream generative quality, bridging the gap between reconstruction-focused evaluation and actual generation performance. We hope this research can provide the community with a new direction in designing effective tokenizers for generation models. 
\renewcommand{\thetable}{{\Alph{table}}}
\renewcommand{\thefigure}{{\Alph{figure}}}
\renewcommand{\thesection}{{\Alph{section}}}
\setcounter{figure}{0}   
\setcounter{table}{0}   
\setcounter{section}{0}
\definecolor{ye}{HTML}{b8aeeb}
\definecolor{NoneColor}{HTML}{f57c6e}
\definecolor{HalfColor}{HTML}{84c3b7}
\definecolor{ZeroColor}{HTML}{b8aeeb}
\section{Appendix}

\subsection{Codebook Size Selection}

\begin{figure*}[]
\centering\includegraphics[width=0.95\linewidth]{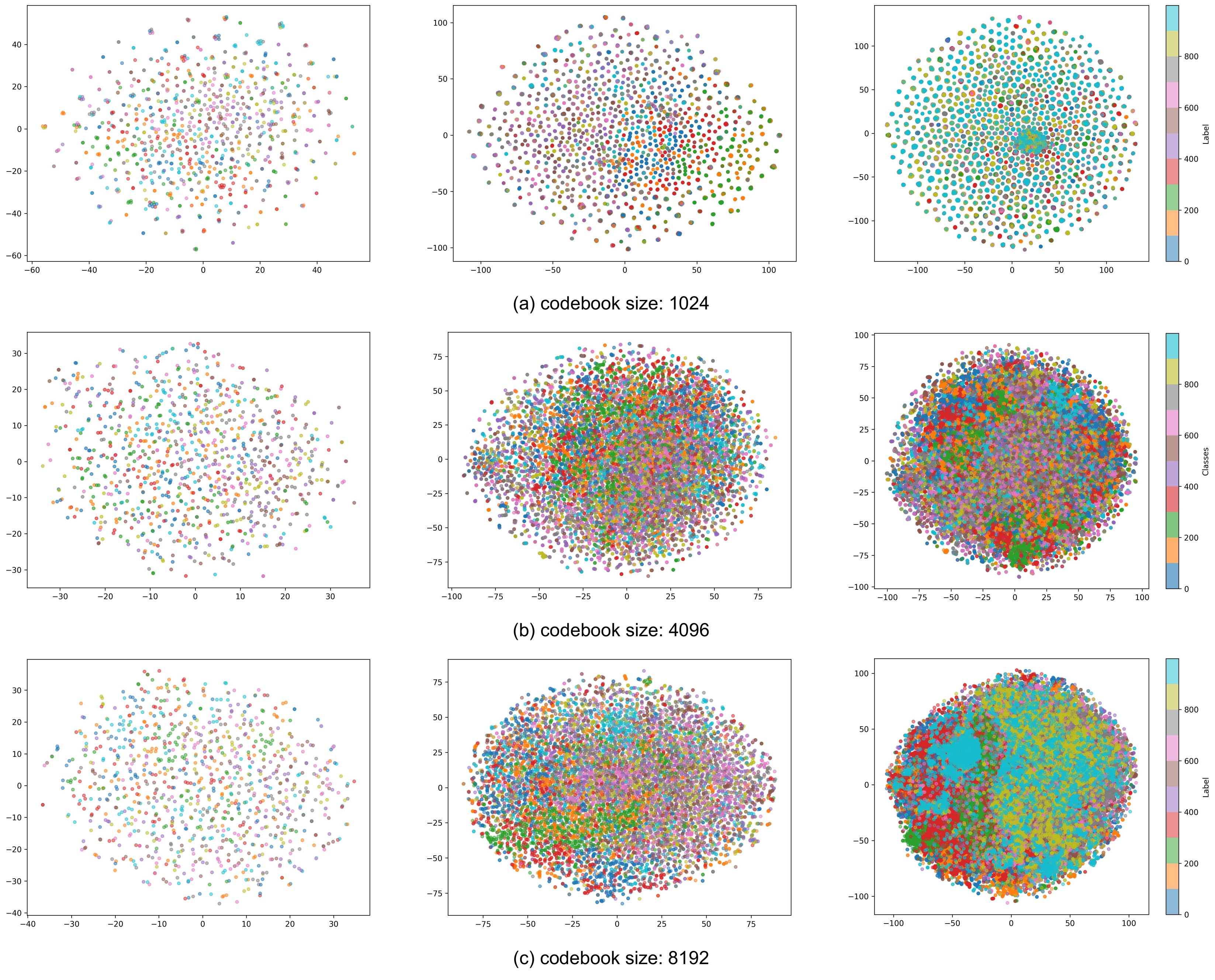}
    \vspace{-0.1in}
    \caption{T-SNE visualization of latent space in XQGAN with varying codebook sizes setting: (a) 1024, (b) 4096, and (c) 8192. Each subfigure presents embeddings derived from (left) 1,000, (middle) 10,000, and (right) 50,000 samples from the ImageNet validation set. Compared to larger codebook sizes, XQGAN-1024 fails to maintain a well-structured latent space, leading to increased fragmentation and reduced robustness.}
    \label{fig:tsne-base}
\end{figure*}

As described in ablation, we initialize our tokenizer with XQGAN-8192 \cite{li2024xq}. Motivated by insights from \cite{yu2024randomized,weber2024maskbit} and our own benchmarking, we aim to reduce the codebook size for a more compact representation while preserving high reconstruction fidelity and generative quality. However, as shown in \cref{fig:tsne-base}, the latent space of images in XQGAN-1024 appears highly fragmented, resulting in notable robustness discrepancies compared to tokenizers with larger codebooks, such as XQGAN-8192 and XQGAN-16384.

To better understand this, we analyze DINO features on ImageNet and apply k-means clustering to feature embeddings. As shown in \cref{tab:elbow-method}, the results of the clustering of k-means, evaluated using the elbow method, indicate decreasing improvements in the Sum of Squared Errors (SSE) as the number of clusters increases beyond 4096. The reduction in SSE slows significantly at this point, suggesting that further increasing the number of clusters yields only marginal benefits. Based on this observation, we select $K = 4096$ as the codebook size for our tokenizer.

\begin{table}
    \centering
    \resizebox{0.45\textwidth}{!}{ 
    \begin{tabular}{c|cccccc}
        Cluster Number & 512 & 1024 & 2048 & 4096 & 8192 & 16384 \\
        \hline
        SSE. & 2250$_{\textcolor{NoneColor}{0}}$ & 1637$_{\textcolor{NoneColor}{-613}}$ & 1253$_{\textcolor{NoneColor}{-384}}$ & 928$_{\textcolor{NoneColor}{-325}}$ & 611$_{\textcolor{NoneColor}{-317}}$ & 473$_{\textcolor{NoneColor}{-138}}$ \\
    \end{tabular}
    }
    \caption{K-means clustering analysis of DINO features in ImageNet validation set. SSE. denotes as the Sum of Squared Error. The \textcolor{NoneColor}{subscript values} represent the difference in SSE. relative to the previous cluster number, indicating the reduction in error as the number of clusters increases.}
    \vspace{-0.2in}
    \label{tab:elbow-method}
\end{table}

\subsection{Loss Function.}
The RobustTok is trained with composite losses including reconstruction loss $\mathcal{L}_{recon}$, vector quantization loss $\mathcal{L}_{VQ}$ \cite{esser2021taming}, adverserial loss $\mathcal{L}_{ad}$ \cite{karras2019style}, Perceptual loss $\mathcal{L}_{P}$ \cite{ledig2017photo}, and semantic loss $\mathcal{L}_{clip}$ \cite{li2024imagefolder}:
\begin{equation}
    \mathcal{L}=\lambda_{rec}\mathcal{L}_{rec} + 
    \lambda_{VQ}\mathcal{L}_{VQ} + \lambda_{ad}\mathcal{L}_{ad} + \lambda_{P}\mathcal{L}_{P} + \lambda_{sem}\mathcal{L}_{sem}.
\end{equation}
Specifically, the reconstruction loss measures the $L_2$ distance between the reconstructed image and the ground truth; vector quantization loss encourages the encoded features and its aligned codebook vectors; adversarial loss ensures that the generated images are indistinguishable from real ones; perceptual loss compares high-level feature representations to capture structural differences; and semantic loss performs semantic regularization between semantic tokens and the pre-trained DINOv2 \cite{oquab2023dinov2} features.

\begin{figure*}
  \centering
  \begin{subfigure}{0.45\linewidth}
    \includegraphics[width=\linewidth]{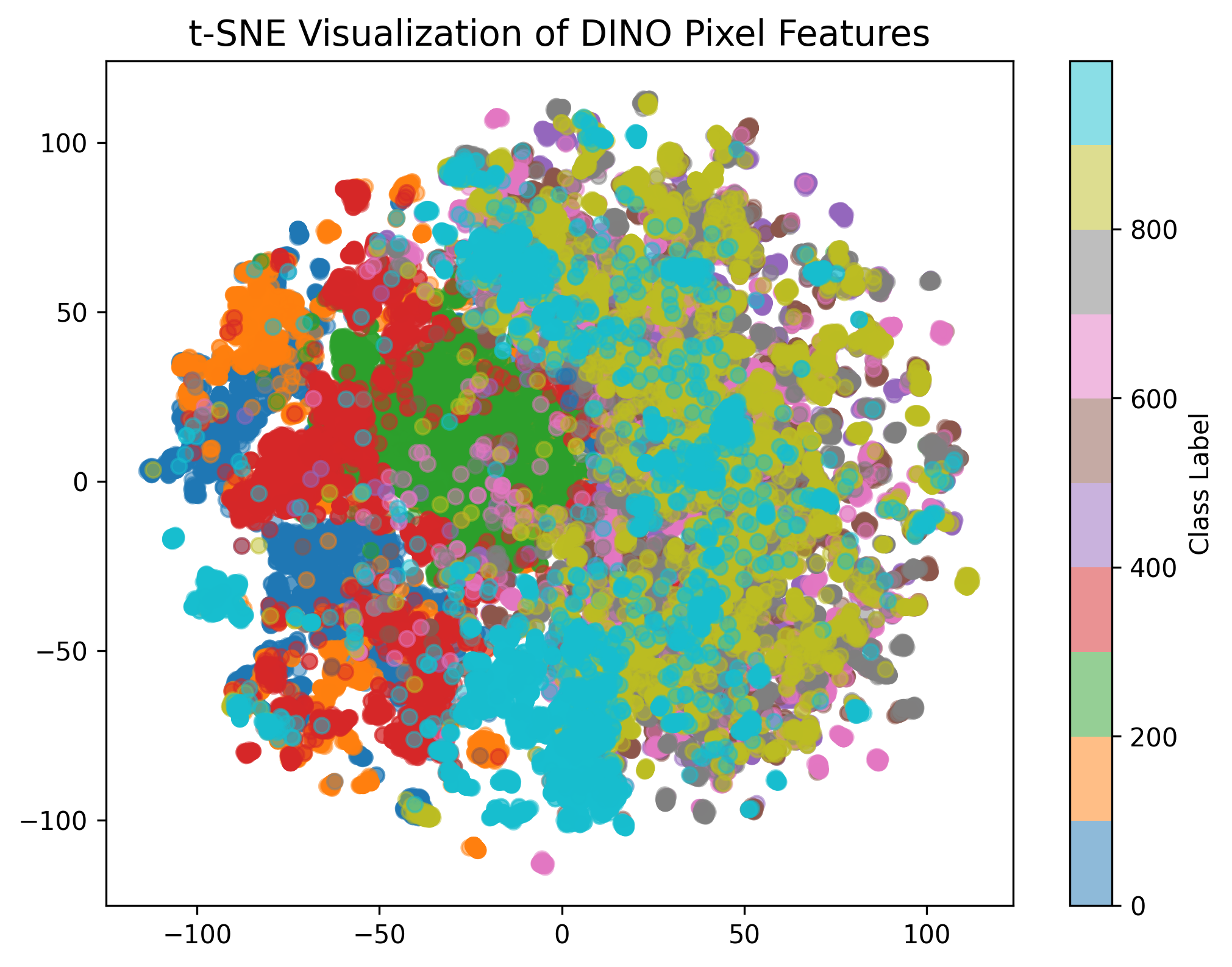}
    \caption{T-SNE visualization of DINO Pixel features.}
    \label{fig:short-a}
  \end{subfigure}
  \hfill
  \begin{subfigure}{0.45\linewidth}
    \includegraphics[width=\linewidth]{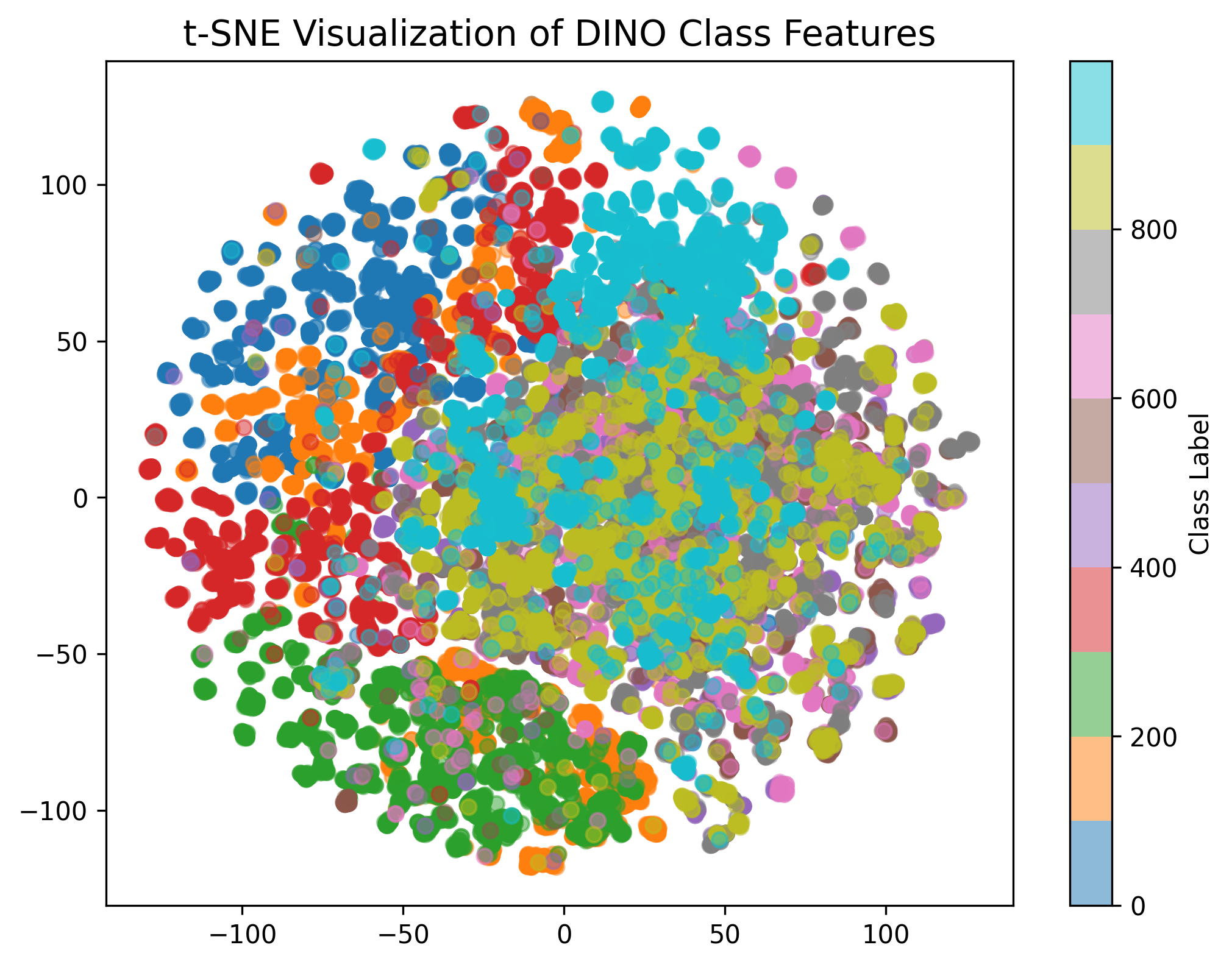}
    \caption{T-SNE visualization of DINO Class features.}
    \label{fig:short-b}
  \end{subfigure}
  \caption{Visualization of DINO features in ImageNet Validation Set.}
  \label{fig:short}
\end{figure*}

\paragraph{DINO supervision.}
As shown in \cref{fig:short}, we visualize the means of DINO pixel features (\cref{fig:short-a}) and DINO class features (\cref{fig:short-b}). We observe that DINO class features exhibit a more structured representation compared to pixel-level features, which appear to be more scattered. Since the purpose of DINO features in our model is to provide supervision, the structured nature of class features makes them a more suitable choice to guide the learning process.

\begin{figure}
    \centering
    \includegraphics[width=1.0\linewidth]{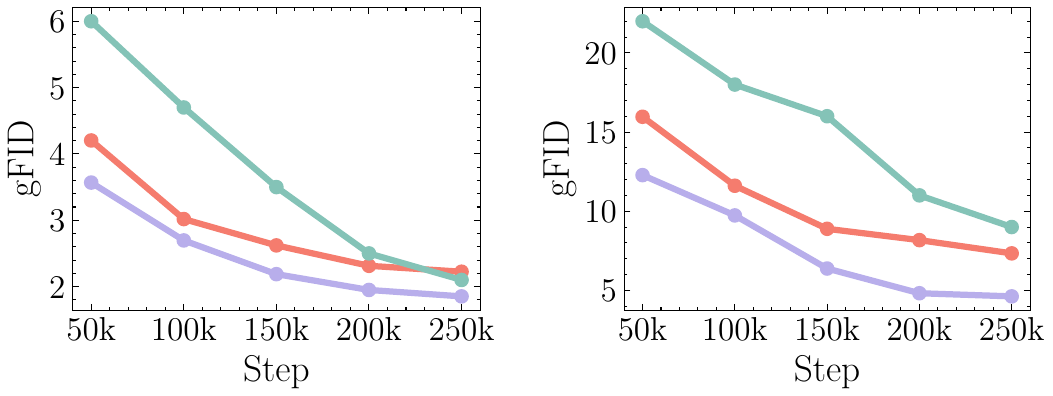}
    \caption{Visualization of gFID trends for \textcolor{HalfColor}{RAR}, \textcolor{NoneColor}{XQGAN}, and \textcolor{ye}{Ours} with (left) and without (right) CFG.}
    \label{fig:gfid-trend}
\end{figure}

\subsection{RAR Training}

We follow the RAR training setting to validate the performance of our RobustTok. Specifically, as shown in \cref{fig:gfid-trend}, we evaluate RAR, XQGAN(our baseline), and our proposed RobustTok during training. We observe that XQGAN achieves a faster convergence speed and better performance without CFG; however, its final performance, with a gFID of 2.22 under classifier-free guidance (CFG), remains suboptimal compared to RAR. Our RobustTok, inheriting the structural advantages of the semantic tokenizer while incorporating a robust latent space, not only achieves faster convergence but also outperforms both XQGAN and vanilla RAR in final generative quality, demonstrating its effectiveness in preserving semantic consistency and enhancing feature representation. This highlights a promising direction for designing more robust training schemes to further improve generative performance.

\begin{figure*}
\centering
\includegraphics[width=\linewidth]{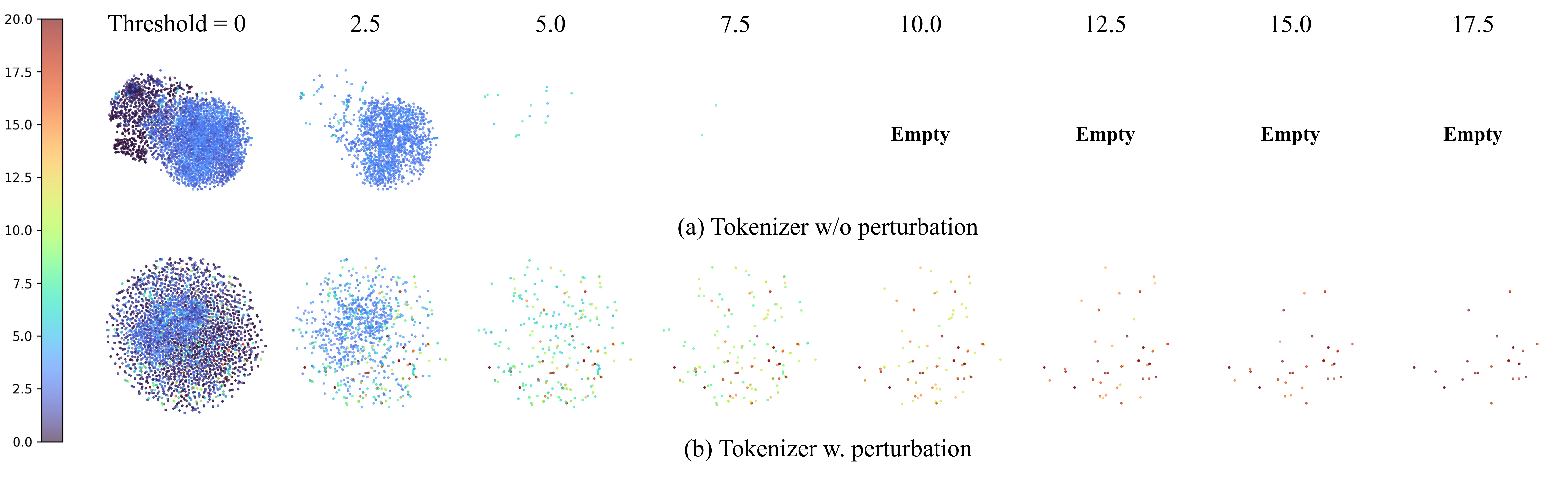}
    \caption{Detailed t-SNE visualization of latent space of tokenizer training with and without our proposed latent perturbation.}
    \vspace{-0.1in}
    \label{fig:tsne-intuition}
\end{figure*}

\begin{table*}
\begin{center}
\renewcommand{\arraystretch}{1.2} 
\setlength{\tabcolsep}{8pt} 
\scalebox{0.85}{
\begin{tabular}{c|
p{3cm}<{\centering}
p{5cm}<{\centering}|
p{1.5cm}<{\centering}
p{1.5cm}<{\centering}|
p{1.5cm}<{\centering}
p{1.8cm}<{\centering}}
\hline
\hline
\multirow{2}{*}{Codebook Size} & \multirow{2}{*}{Method} & \multirow{2}{*}{Tokenizer Type} & \multicolumn{2}{c}{Tokenizer} & \multicolumn{2}{c}{Generator} \\
    \cline{4-7}
    & & & rFID & pFID & gFID & gFID (CFG) \\

\hline
\multirow{4}*{16384}& VQGAN-16384 \cite{esser2021taming} & Non-semantic & 4.50 & 18.18 & 20.89 & 6.23 \\
~ & LlamaGen \cite{sun2024autoregressive} & Non-semantic & 2.19 & 13.12 & 8.61 & 4.40 \\
& IBQ-16384 \cite{shi2024taming} & Non-semantic & 1.41 & 16.35 & 21.57 &  5.53 \\
~ & VQGAN-LC \cite{zhu2024scaling} & Trainable Projector & 3.27 & 16.78 & 17.55 & 5.50 \\
\hline
\multirow{3}*{8192} & IBQ-8192 \cite{shi2024taming} & Non-semantic & 1.87 & 19.62& 21.05 & 5.41 \\
~ & TiTok \cite{yu2024imageworth32tokens} & Semantic &  1.03 & 3.55 & 14.51 & 4.47 \\
~ & XQGAN-8192 \cite{li2024xq} & Semantic & \textbf{0.81} & 7.91 & 14.64 & 4.48 \\
\hline
\multirow{2}*{4096} & XQGAN-4096 \cite{li2024xq} & Semantic & 0.91 & 6.98 & 7.90 & 4.13 \\
 & \cellcolor{blue!8}RobustTok (Ours) & \cellcolor{blue!8}Semantic + Robust & \cellcolor{blue!8}1.02 & \cellcolor{blue!8}\textbf{2.28} & \cellcolor{blue!8}\textbf{6.47} & \cellcolor{blue!8}\textbf{3.51} \\

\hline
\multirow{2}*{1024} &  MaskGIT \cite{chang2022maskgitmaskedgenerativeimage} & Non-Semantic & 2.28 & 4.20 & 12.37 & 3.60\\
& IBQ-1024 \cite{shi2024taming} & Index Non-Semantic & 2.24 & 6.37 & 23.89 & 5.53\\

\hline
\hline
\end{tabular}
}
\caption{Tokenizer benchmarking for LlamaGen-L. All metrics, \ie, rFID, pFID and gFID, are the smaller the better.}
\label{tab:token-setting-l}

\end{center}    
\end{table*}

\begin{figure*}
    \centering
    \begin{subfigure}{0.48\linewidth}
        \centering
        \includegraphics[width=\linewidth]{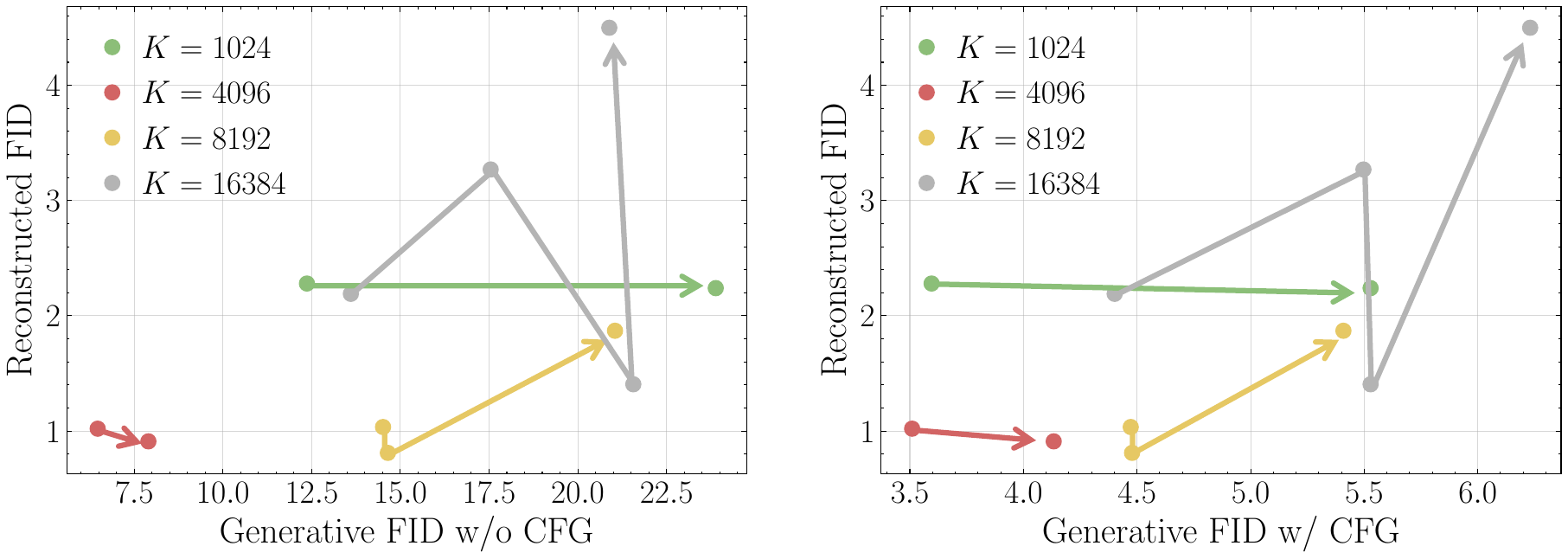}
        \caption{\textbf{rFID} v.s gFID with and without CFG.}
        \label{fig:pfid-LlamaGen-L-recon}
    \end{subfigure}
    \hfill
    \begin{subfigure}{0.48\linewidth}
        \centering
        \includegraphics[width=\linewidth]{fig/Llama-B-perturb.pdf}
        \caption{\textbf{pFID} vs. gFID with and without CFG.}
        \label{fig:pfid-LlamaGen-L-pert}
    \end{subfigure}
    \caption{Comparison of reconstructed FID relation to generative FID with perturbed FID relation to generative FID. All generators follow LlamaGen-L training setting. K denotes as codebook size.}
    \vspace{-0.01in}
    \label{fig:pfid-LlamaGen-L}
\end{figure*}

\begin{figure*}
\centering\includegraphics[width=0.95\linewidth]{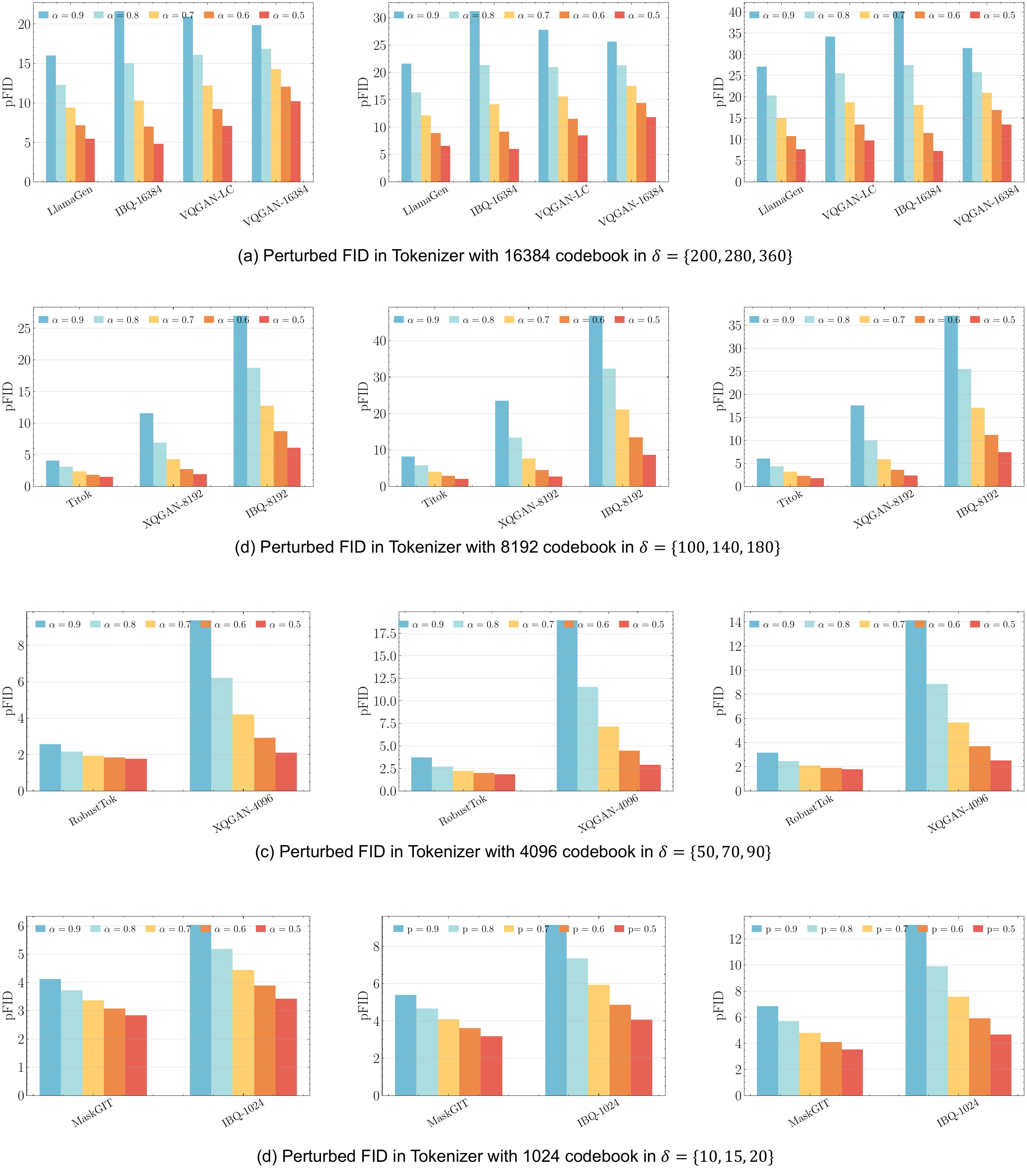}
    \caption{qualitative analysis of tokenizers in our latent perturbation.}
    \label{fig:pfid}
\end{figure*}

\subsection{Latent Perturbation v.s. Other Noises}
To avoid potential misunderstanding, we aim to discuss the difference between our proposed latent perturbation and other noises used in generative models.

\begin{itemize}
    \item \textbf{Latent perturbation}: Latent perturbation is a \textbf{random} noise manually added to the latent space based on the pattern we observed during the real sampling errors. Specifically, it is added in a cluster-based manner enlarging the decision boundary and zero-shot generalization during inference.
    \item \textbf{Diffusion noise}: Diffusion noise is a \textbf{scheduled} noise added to enable the reverse process using a diffusion sampler. It follows a pre-defined schedule to systematically disrupt the latent space.
    \item \textbf{Gaussian noise in VAE}: VAE's reparameterization employs a gaussian noise to decompose the mean value and randomness of the distribution to enable the gradient backpropagation. 

\end{itemize}
\clearpage
{
    \small
    \bibliographystyle{ieeenat_fullname}
    \bibliography{main}

\begin{thebibliography}{92}
\providecommand{\natexlab}[1]{#1}
\providecommand{\url}[1]{\texttt{#1}}
\expandafter\ifx\csname urlstyle\endcsname\relax
  \providecommand{\doi}[1]{doi: #1}\else
  \providecommand{\doi}{doi: \begingroup \urlstyle{rm}\Url}\fi

\bibitem[Achiam et~al.(2023)Achiam, Adler, Agarwal, Ahmad, Akkaya, Aleman, Almeida, Altenschmidt, Altman, Anadkat, et~al.]{achiam2023gpt}
Josh Achiam, Steven Adler, Sandhini Agarwal, Lama Ahmad, Ilge Akkaya, Florencia~Leoni Aleman, Diogo Almeida, Janko Altenschmidt, Sam Altman, Shyamal Anadkat, et~al.
\newblock Gpt-4 technical report.
\newblock \emph{arXiv preprint arXiv:2303.08774}, 2023.

\bibitem[Bachmann et~al.(2025)Bachmann, Allardice, Mizrahi, Fini, Kar, Amirloo, El-Nouby, Zamir, and Dehghan]{bachmann2025flextok}
Roman Bachmann, Jesse Allardice, David Mizrahi, Enrico Fini, O{\u{g}}uzhan~Fatih Kar, Elmira Amirloo, Alaaeldin El-Nouby, Amir Zamir, and Afshin Dehghan.
\newblock Flextok: Resampling images into 1d token sequences of flexible length.
\newblock \emph{arXiv preprint arXiv:2502.13967}, 2025.

\bibitem[Bengio et~al.(2015)Bengio, Vinyals, Jaitly, and Shazeer]{bengio2015scheduled}
Samy Bengio, Oriol Vinyals, Navdeep Jaitly, and Noam Shazeer.
\newblock Scheduled sampling for sequence prediction with recurrent neural networks.
\newblock \emph{Advances in neural information processing systems}, 28, 2015.

\bibitem[Chang et~al.(2022)Chang, Zhang, Jiang, Liu, and Freeman]{chang2022maskgitmaskedgenerativeimage}
Huiwen Chang, Han Zhang, Lu Jiang, Ce Liu, and William~T. Freeman.
\newblock Maskgit: Masked generative image transformer, 2022.

\bibitem[Chen et~al.(2024{\natexlab{a}})Chen, Han, Misra, Li, Hu, Zou, Sugiyama, Wang, and Raj]{chen2024slight}
Hao Chen, Yujin Han, Diganta Misra, Xiang Li, Kai Hu, Difan Zou, Masashi Sugiyama, Jindong Wang, and Bhiksha Raj.
\newblock Slight corruption in pre-training data makes better diffusion models.
\newblock \emph{arXiv preprint arXiv:2405.20494}, 2024{\natexlab{a}}.

\bibitem[Chen et~al.(2024{\natexlab{b}})Chen, Wang, Li, Sun, Chen, Liu, Wang, Raj, Liu, and Barsoum]{chen2024softvq}
Hao Chen, Ze Wang, Xiang Li, Ximeng Sun, Fangyi Chen, Jiang Liu, Jindong Wang, Bhiksha Raj, Zicheng Liu, and Emad Barsoum.
\newblock Softvq-vae: Efficient 1-dimensional continuous tokenizer.
\newblock \emph{arXiv preprint arXiv:2412.10958}, 2024{\natexlab{b}}.

\bibitem[Chen et~al.(2025)Chen, Han, Chen, Li, Wang, Wang, Wang, Liu, Zou, and Raj]{chen2025masked}
Hao Chen, Yujin Han, Fangyi Chen, Xiang Li, Yidong Wang, Jindong Wang, Ze Wang, Zicheng Liu, Difan Zou, and Bhiksha Raj.
\newblock Masked autoencoders are effective tokenizers for diffusion models.
\newblock \emph{arXiv preprint arXiv:2502.03444}, 2025.

\bibitem[Chiu et~al.(2018)Chiu, Sainath, Wu, Prabhavalkar, Nguyen, Chen, Kannan, Weiss, Rao, Gonina, et~al.]{chiu2018state}
Chung-Cheng Chiu, Tara~N Sainath, Yonghui Wu, Rohit Prabhavalkar, Patrick Nguyen, Zhifeng Chen, Anjuli Kannan, Ron~J Weiss, Kanishka Rao, Ekaterina Gonina, et~al.
\newblock State-of-the-art speech recognition with sequence-to-sequence models.
\newblock In \emph{2018 IEEE international conference on acoustics, speech and signal processing (ICASSP)}, pages 4774--4778. IEEE, 2018.

\bibitem[Darcet et~al.(2023)Darcet, Oquab, Mairal, and Bojanowski]{darcet2023vitneedreg}
Timothée Darcet, Maxime Oquab, Julien Mairal, and Piotr Bojanowski.
\newblock Vision transformers need registers, 2023.

\bibitem[Deng et~al.(2009)Deng, Dong, Socher, Li, Li, and Fei-Fei]{deng2009imagenet}
Jia Deng, Wei Dong, Richard Socher, Li-Jia Li, Kai Li, and Li Fei-Fei.
\newblock Imagenet: A large-scale hierarchical image database.
\newblock In \emph{2009 IEEE conference on computer vision and pattern recognition}, pages 248--255. Ieee, 2009.

\bibitem[Dhariwal and Nichol(2021)]{dhariwal2021diffusionmodelsbeatgans}
Prafulla Dhariwal and Alex Nichol.
\newblock Diffusion models beat gans on image synthesis, 2021.

\bibitem[Dong et~al.(2023)Dong, Bao, Zhang, Chen, Zhang, Yuan, Chen, Wen, Yu, and Guo]{dong2023peco}
Xiaoyi Dong, Jianmin Bao, Ting Zhang, Dongdong Chen, Weiming Zhang, Lu Yuan, Dong Chen, Fang Wen, Nenghai Yu, and Baining Guo.
\newblock Peco: Perceptual codebook for bert pre-training of vision transformers.
\newblock In \emph{Proceedings of the AAAI Conference on Artificial Intelligence}, pages 552--560, 2023.

\bibitem[Dosovitskiy et~al.(2020)Dosovitskiy, Beyer, Kolesnikov, Weissenborn, Zhai, Unterthiner, Dehghani, Minderer, Heigold, Gelly, et~al.]{dosovitskiy2020image}
Alexey Dosovitskiy, Lucas Beyer, Alexander Kolesnikov, Dirk Weissenborn, Xiaohua Zhai, Thomas Unterthiner, Mostafa Dehghani, Matthias Minderer, Georg Heigold, Sylvain Gelly, et~al.
\newblock An image is worth 16x16 words: Transformers for image recognition at scale.
\newblock \emph{arXiv preprint arXiv:2010.11929}, 2020.

\bibitem[Dosovitskiy et~al.(2021)Dosovitskiy, Beyer, Kolesnikov, Weissenborn, Zhai, Unterthiner, Dehghani, Minderer, Heigold, Gelly, Uszkoreit, and Houlsby]{dosovitskiy2021imageworth16x16words}
Alexey Dosovitskiy, Lucas Beyer, Alexander Kolesnikov, Dirk Weissenborn, Xiaohua Zhai, Thomas Unterthiner, Mostafa Dehghani, Matthias Minderer, Georg Heigold, Sylvain Gelly, Jakob Uszkoreit, and Neil Houlsby.
\newblock An image is worth 16x16 words: Transformers for image recognition at scale, 2021.

\bibitem[Esser et~al.(2021)Esser, Rombach, and Ommer]{esser2021taming}
Patrick Esser, Robin Rombach, and Bjorn Ommer.
\newblock Taming transformers for high-resolution image synthesis.
\newblock In \emph{Proceedings of the IEEE/CVF conference on computer vision and pattern recognition}, pages 12873--12883, 2021.

\bibitem[Fan et~al.(2024)Fan, Li, Qin, Li, Sun, Rubinstein, Sun, He, and Tian]{fan2024fluid}
Lijie Fan, Tianhong Li, Siyang Qin, Yuanzhen Li, Chen Sun, Michael Rubinstein, Deqing Sun, Kaiming He, and Yonglong Tian.
\newblock Fluid: Scaling autoregressive text-to-image generative models with continuous tokens.
\newblock \emph{arXiv preprint arXiv:2410.13863}, 2024.

\bibitem[Han et~al.(2024)Han, Liu, Jiang, Yan, Zhang, Yuan, Peng, and Liu]{han2024infinity}
Jian Han, Jinlai Liu, Yi Jiang, Bin Yan, Yuqi Zhang, Zehuan Yuan, Bingyue Peng, and Xiaobing Liu.
\newblock Infinity: Scaling bitwise autoregressive modeling for high-resolution image synthesis.
\newblock \emph{arXiv preprint arXiv:2412.04431}, 2024.

\bibitem[He et~al.(2022)He, Chen, Xie, Li, Doll{\'a}r, and Girshick]{he2022masked}
Kaiming He, Xinlei Chen, Saining Xie, Yanghao Li, Piotr Doll{\'a}r, and Ross Girshick.
\newblock Masked autoencoders are scalable vision learners.
\newblock In \emph{Proceedings of the IEEE/CVF conference on computer vision and pattern recognition}, pages 16000--16009, 2022.

\bibitem[He et~al.(2024)He, Fu, Liu, Wang, Xiao, Shu, Wang, Zhang, Yu, Li, et~al.]{he2024mars}
Wanggui He, Siming Fu, Mushui Liu, Xierui Wang, Wenyi Xiao, Fangxun Shu, Yi Wang, Lei Zhang, Zhelun Yu, Haoyuan Li, et~al.
\newblock Mars: Mixture of auto-regressive models for fine-grained text-to-image synthesis.
\newblock \emph{arXiv preprint arXiv:2407.07614}, 2024.

\bibitem[Heusel et~al.(2017)Heusel, Ramsauer, Unterthiner, Nessler, and Hochreiter]{heusel2017gans}
Martin Heusel, Hubert Ramsauer, Thomas Unterthiner, Bernhard Nessler, and Sepp Hochreiter.
\newblock Gans trained by a two time-scale update rule converge to a local nash equilibrium.
\newblock \emph{Advances in Neural Information Processing Systems}, 30, 2017.

\bibitem[Hinton and Salakhutdinov(2006)]{hinton2006reducing}
Geoffrey~E Hinton and Ruslan~R Salakhutdinov.
\newblock Reducing the dimensionality of data with neural networks.
\newblock \emph{science}, 313\penalty0 (5786):\penalty0 504--507, 2006.

\bibitem[Ho and Salimans(2022)]{ho2022classifier}
Jonathan Ho and Tim Salimans.
\newblock Classifier-free diffusion guidance.
\newblock \emph{arXiv preprint arXiv:2207.12598}, 2022.

\bibitem[Huang et~al.(2023)Huang, Mao, Chen, and Zhang]{huang2023towards}
Mengqi Huang, Zhendong Mao, Zhuowei Chen, and Yongdong Zhang.
\newblock Towards accurate image coding: Improved autoregressive image generation with dynamic vector quantization.
\newblock In \emph{Proceedings of the IEEE/CVF Conference on Computer Vision and Pattern Recognition}, pages 22596--22605, 2023.

\bibitem[Karras et~al.(2019)Karras, Laine, and Aila]{karras2019style}
Tero Karras, Samuli Laine, and Timo Aila.
\newblock A style-based generator architecture for generative adversarial networks.
\newblock In \emph{Proceedings of the IEEE/CVF conference on computer vision and pattern recognition}, pages 4401--4410, 2019.

\bibitem[Kim et~al.(2025)Kim, He, Yu, Yang, Shen, Kwak, and Chen]{kim2025democratizing}
Dongwon Kim, Ju He, Qihang Yu, Chenglin Yang, Xiaohui Shen, Suha Kwak, and Liang-Chieh Chen.
\newblock Democratizing text-to-image masked generative models with compact text-aware one-dimensional tokens.
\newblock \emph{arXiv preprint arXiv:2501.07730}, 2025.

\bibitem[Lamb et~al.(2016)Lamb, ALIAS PARTH~GOYAL, Zhang, Zhang, Courville, and Bengio]{lamb2016professor}
Alex~M Lamb, Anirudh~Goyal ALIAS PARTH~GOYAL, Ying Zhang, Saizheng Zhang, Aaron~C Courville, and Yoshua Bengio.
\newblock Professor forcing: A new algorithm for training recurrent networks.
\newblock \emph{Advances in neural information processing systems}, 29, 2016.

\bibitem[Ledig et~al.(2017)Ledig, Theis, Husz{\'a}r, Caballero, Cunningham, Acosta, Aitken, Tejani, Totz, Wang, et~al.]{ledig2017photo}
Christian Ledig, Lucas Theis, Ferenc Husz{\'a}r, Jose Caballero, Andrew Cunningham, Alejandro Acosta, Andrew Aitken, Alykhan Tejani, Johannes Totz, Zehan Wang, et~al.
\newblock Photo-realistic single image super-resolution using a generative adversarial network.
\newblock In \emph{Proceedings of the IEEE conference on computer vision and pattern recognition}, pages 4681--4690, 2017.

\bibitem[Lee et~al.(2022{\natexlab{a}})Lee, Kim, Kim, Cho, and Han]{lee2022autoregressive}
Doyup Lee, Chiheon Kim, Saehoon Kim, Minsu Cho, and Wook-Shin Han.
\newblock Autoregressive image generation using residual quantization.
\newblock In \emph{Proceedings of the IEEE/CVF Conference on Computer Vision and Pattern Recognition}, pages 11523--11532, 2022{\natexlab{a}}.

\bibitem[Lee et~al.(2022{\natexlab{b}})Lee, Kim, Kim, Cho, and Han]{lee2022autoregressiveimagegenerationusing}
Doyup Lee, Chiheon Kim, Saehoon Kim, Minsu Cho, and Wook-Shin Han.
\newblock Autoregressive image generation using residual quantization, 2022{\natexlab{b}}.

\bibitem[Li et~al.(2024{\natexlab{a}})Li, Yang, Wang, Qiu, Chou, Li, and Li]{li2024scalable}
Haopeng Li, Jinyue Yang, Kexin Wang, Xuerui Qiu, Yuhong Chou, Xin Li, and Guoqi Li.
\newblock Scalable autoregressive image generation with mamba.
\newblock \emph{arXiv preprint arXiv:2408.12245}, 2024{\natexlab{a}}.

\bibitem[Li et~al.(2024{\natexlab{b}})Li, Katabi, and He]{li2024returnunconditionalgenerationselfsupervised}
Tianhong Li, Dina Katabi, and Kaiming He.
\newblock Return of unconditional generation: A self-supervised representation generation method, 2024{\natexlab{b}}.

\bibitem[Li et~al.(2024{\natexlab{c}})Li, Tian, Li, Deng, and He]{li2024autoregressiveimagegenerationvector}
Tianhong Li, Yonglong Tian, He Li, Mingyang Deng, and Kaiming He.
\newblock Autoregressive image generation without vector quantization, 2024{\natexlab{c}}.

\bibitem[Li et~al.(2023{\natexlab{a}})Li, Wang, Xu, Li, Raj, and Lu]{li2023robust}
Xiang Li, Jinglu Wang, Xiaohao Xu, Xiao Li, Bhiksha Raj, and Yan Lu.
\newblock Robust referring video object segmentation with cyclic structural consensus.
\newblock In \emph{Proceedings of the IEEE/CVF International Conference on Computer Vision}, pages 22236--22245, 2023{\natexlab{a}}.

\bibitem[Li et~al.(2023{\natexlab{b}})Li, Wang, Xu, Yang, Yang, Zhao, Singh, and Raj]{li2023towards}
Xiang Li, Jinglu Wang, Xiaohao Xu, Muqiao Yang, Fan Yang, Yizhou Zhao, Rita Singh, and Bhiksha Raj.
\newblock Towards noise-tolerant speech-referring video object segmentation: Bridging speech and text.
\newblock In \emph{Proceedings of the 2023 Conference on Empirical Methods in Natural Language Processing}, pages 2283--2296, 2023{\natexlab{b}}.

\bibitem[Li et~al.(2024{\natexlab{d}})Li, Qiu, Chen, Kuen, Gu, Raj, and Lin]{li2024imagefolder}
Xiang Li, Kai Qiu, Hao Chen, Jason Kuen, Jiuxiang Gu, Bhiksha Raj, and Zhe Lin.
\newblock Imagefolder: Autoregressive image generation with folded tokens.
\newblock \emph{arXiv preprint arXiv:2410.01756}, 2024{\natexlab{d}}.

\bibitem[Li et~al.(2024{\natexlab{e}})Li, Qiu, Chen, Kuen, Gu, Wang, Lin, and Raj]{li2024xq}
Xiang Li, Kai Qiu, Hao Chen, Jason Kuen, Jiuxiang Gu, Jindong Wang, Zhe Lin, and Bhiksha Raj.
\newblock Xq-gan: An open-source image tokenization framework for autoregressive generation.
\newblock \emph{arXiv preprint arXiv:2412.01762}, 2024{\natexlab{e}}.

\bibitem[Li et~al.(2024{\natexlab{f}})Li, Qiu, Chen, Kuen, Lin, Singh, and Raj]{li2024controlvar}
Xiang Li, Kai Qiu, Hao Chen, Jason Kuen, Zhe Lin, Rita Singh, and Bhiksha Raj.
\newblock Controlvar: Exploring controllable visual autoregressive modeling.
\newblock \emph{arXiv preprint arXiv:2406.09750}, 2024{\natexlab{f}}.

\bibitem[Li et~al.(2024{\natexlab{g}})Li, Qiu, Wang, Xu, Singh, Yamazaki, Chen, Huang, and Raj]{li2024r}
Xiang Li, Kai Qiu, Jinglu Wang, Xiaohao Xu, Rita Singh, Kashu Yamazaki, Hao Chen, Xiaonan Huang, and Bhiksha Raj.
\newblock R 2-bench: Benchmarking the robustness of referring perception models under perturbations.
\newblock In \emph{European Conference on Computer Vision}, pages 211--230. Springer, 2024{\natexlab{g}}.

\bibitem[Li et~al.(2024{\natexlab{h}})Li, Wang, Xu, Peng, Singh, Lu, and Raj]{li2024qdformer}
Xiang Li, Jinglu Wang, Xiaohao Xu, Xiulian Peng, Rita Singh, Yan Lu, and Bhiksha Raj.
\newblock Qdformer: towards robust audiovisual segmentation in complex environments with quantization-based semantic decomposition.
\newblock In \emph{Proceedings of the IEEE/CVF Conference on Computer Vision and Pattern Recognition}, pages 3402--3413, 2024{\natexlab{h}}.

\bibitem[Luo et~al.(2024)Luo, Shi, Ge, Yang, Wang, and Shan]{luo2024open}
Zhuoyan Luo, Fengyuan Shi, Yixiao Ge, Yujiu Yang, Limin Wang, and Ying Shan.
\newblock Open-magvit2: An open-source project toward democratizing auto-regressive visual generation.
\newblock \emph{arXiv preprint arXiv:2409.04410}, 2024.

\bibitem[Mentzer et~al.(2023)Mentzer, Minnen, Agustsson, and Tschannen]{mentzer2023finite}
Fabian Mentzer, David Minnen, Eirikur Agustsson, and Michael Tschannen.
\newblock Finite scalar quantization: Vq-vae made simple, 2023.

\bibitem[Miwa et~al.(2025)Miwa, Sasaki, Arai, Takahashi, and Yamaguchi]{miwa2025one}
Keita Miwa, Kento Sasaki, Hidehisa Arai, Tsubasa Takahashi, and Yu Yamaguchi.
\newblock One-d-piece: Image tokenizer meets quality-controllable compression.
\newblock \emph{arXiv e-prints}, pages arXiv--2501, 2025.

\bibitem[Mizrahi et~al.(2024)Mizrahi, Bachmann, Kar, Yeo, Gao, Dehghan, and Zamir]{mizrahi20244m}
David Mizrahi, Roman Bachmann, Oguzhan Kar, Teresa Yeo, Mingfei Gao, Afshin Dehghan, and Amir Zamir.
\newblock 4m: Massively multimodal masked modeling.
\newblock \emph{Advances in Neural Information Processing Systems}, 36, 2024.

\bibitem[Nichol and Dhariwal(2021)]{nichol2021improveddenoisingdiffusionprobabilistic}
Alex Nichol and Prafulla Dhariwal.
\newblock Improved denoising diffusion probabilistic models, 2021.

\bibitem[Ning et~al.()Ning, Li, Zhang, Geng, Dai, He, and Hu]{ning2301all}
J Ning, C Li, Z Zhang, Z Geng, Q Dai, K He, and H Hu.
\newblock All in tokens: Unifying output space of visual tasks via soft token. arxiv 2023.
\newblock \emph{arXiv preprint arXiv:2301.02229}.

\bibitem[Oquab et~al.(2023)Oquab, Darcet, Moutakanni, Vo, Szafraniec, Khalidov, Fernandez, Haziza, Massa, El-Nouby, Howes, Huang, Xu, Sharma, Li, Galuba, Rabbat, Assran, Ballas, Synnaeve, Misra, Jegou, Mairal, Labatut, Joulin, and Bojanowski]{oquab2023dinov2}
Maxime Oquab, Timothée Darcet, Theo Moutakanni, Huy~V. Vo, Marc Szafraniec, Vasil Khalidov, Pierre Fernandez, Daniel Haziza, Francisco Massa, Alaaeldin El-Nouby, Russell Howes, Po-Yao Huang, Hu Xu, Vasu Sharma, Shang-Wen Li, Wojciech Galuba, Mike Rabbat, Mido Assran, Nicolas Ballas, Gabriel Synnaeve, Ishan Misra, Herve Jegou, Julien Mairal, Patrick Labatut, Armand Joulin, and Piotr Bojanowski.
\newblock Dinov2: Learning robust visual features without supervision, 2023.

\bibitem[Pang et~al.(2024{\natexlab{a}})Pang, Jin, Yang, Lin, Zhu, Tang, Chen, Tay, Lim, Yang, et~al.]{pang2024next}
Yatian Pang, Peng Jin, Shuo Yang, Bin Lin, Bin Zhu, Zhenyu Tang, Liuhan Chen, Francis~EH Tay, Ser-Nam Lim, Harry Yang, et~al.
\newblock Next patch prediction for autoregressive visual generation.
\newblock \emph{arXiv preprint arXiv:2412.15321}, 2024{\natexlab{a}}.

\bibitem[Pang et~al.(2024{\natexlab{b}})Pang, Zhang, Luan, Man, Tan, Zhang, Freeman, and Wang]{pang2024randar}
Ziqi Pang, Tianyuan Zhang, Fujun Luan, Yunze Man, Hao Tan, Kai Zhang, William~T Freeman, and Yu-Xiong Wang.
\newblock Randar: Decoder-only autoregressive visual generation in random orders.
\newblock \emph{arXiv preprint arXiv:2412.01827}, 2024{\natexlab{b}}.

\bibitem[Peebles and Xie(2023)]{peebles2023scalablediffusionmodelstransformers}
William Peebles and Saining Xie.
\newblock Scalable diffusion models with transformers, 2023.

\bibitem[Qiu et~al.(2024)Qiu, Li, Chen, Sun, Wang, Lin, Savvides, and Raj]{qiu2024efficient}
Kai Qiu, Xiang Li, Hao Chen, Jie Sun, Jinglu Wang, Zhe Lin, Marios Savvides, and Bhiksha Raj.
\newblock Efficient autoregressive audio modeling via next-scale prediction.
\newblock \emph{arXiv preprint arXiv:2408.09027}, 2024.

\bibitem[Radford et~al.(2021)Radford, Kim, Hallacy, Ramesh, Goh, Agarwal, Sastry, Askell, Mishkin, Clark, et~al.]{radford2021learning}
Alec Radford, Jong~Wook Kim, Chris Hallacy, Aditya Ramesh, Gabriel Goh, Sandhini Agarwal, Girish Sastry, Amanda Askell, Pamela Mishkin, Jack Clark, et~al.
\newblock Learning transferable visual models from natural language supervision.
\newblock In \emph{International conference on machine learning}, pages 8748--8763. PMLR, 2021.

\bibitem[Razavi et~al.(2019{\natexlab{a}})Razavi, Van~den Oord, and Vinyals]{razavi2019generating}
Ali Razavi, Aaron Van~den Oord, and Oriol Vinyals.
\newblock Generating diverse high-fidelity images with vq-vae-2.
\newblock \emph{Advances in neural information processing systems}, 32, 2019{\natexlab{a}}.

\bibitem[Razavi et~al.(2019{\natexlab{b}})Razavi, van~den Oord, and Vinyals]{razavi2019generatingdiversehighfidelityimages}
Ali Razavi, Aaron van~den Oord, and Oriol Vinyals.
\newblock Generating diverse high-fidelity images with vq-vae-2, 2019{\natexlab{b}}.

\bibitem[Ren et~al.(2024)Ren, Yu, He, Shen, Yuille, and Chen]{ren2024flowar}
Sucheng Ren, Qihang Yu, Ju He, Xiaohui Shen, Alan Yuille, and Liang-Chieh Chen.
\newblock Flowar: Scale-wise autoregressive image generation meets flow matching.
\newblock \emph{arXiv preprint arXiv:2412.15205}, 2024.

\bibitem[Ren et~al.(2025)Ren, Yu, He, Shen, Yuille, and Chen]{ren2025beyond}
Sucheng Ren, Qihang Yu, Ju He, Xiaohui Shen, Alan Yuille, and Liang-Chieh Chen.
\newblock Beyond next-token: Next-x prediction for autoregressive visual generation.
\newblock \emph{arXiv preprint arXiv:2502.20388}, 2025.

\bibitem[Rombach et~al.(2022)Rombach, Blattmann, Lorenz, Esser, and Ommer]{rombach2022highresolutionimagesynthesislatent}
Robin Rombach, Andreas Blattmann, Dominik Lorenz, Patrick Esser, and Björn Ommer.
\newblock High-resolution image synthesis with latent diffusion models, 2022.

\bibitem[Salimans et~al.(2016)Salimans, Goodfellow, Zaremba, Cheung, Radford, and Chen]{salimans2016improved}
Tim Salimans, Ian Goodfellow, Wojciech Zaremba, Vicki Cheung, Alec Radford, and Xi Chen.
\newblock Improved techniques for training gans.
\newblock \emph{Advances in Neural Information Processing Systems}, 29, 2016.

\bibitem[Shi et~al.(2024)Shi, Luo, Ge, Yang, Shan, and Wang]{shi2024taming}
Fengyuan Shi, Zhuoyan Luo, Yixiao Ge, Yujiu Yang, Ying Shan, and Limin Wang.
\newblock Taming scalable visual tokenizer for autoregressive image generation.
\newblock \emph{arXiv preprint arXiv:2412.02692}, 2024.

\bibitem[Shi et~al.(2022)Shi, Wu, Liang, Liu, and Duan]{shi2022divaephotorealisticimagessynthesis}
Jie Shi, Chenfei Wu, Jian Liang, Xiang Liu, and Nan Duan.
\newblock Divae: Photorealistic images synthesis with denoising diffusion decoder, 2022.

\bibitem[Song et~al.(2022)Song, Meng, and Ermon]{song2022denoisingdiffusionimplicitmodels}
Jiaming Song, Chenlin Meng, and Stefano Ermon.
\newblock Denoising diffusion implicit models, 2022.

\bibitem[Sun et~al.(2024)Sun, Jiang, Chen, Zhang, Peng, Luo, and Yuan]{sun2024autoregressive}
Peize Sun, Yi Jiang, Shoufa Chen, Shilong Zhang, Bingyue Peng, Ping Luo, and Zehuan Yuan.
\newblock Autoregressive model beats diffusion: Llama for scalable image generation.
\newblock \emph{arXiv preprint arXiv:2406.06525}, 2024.

\bibitem[Sutton(1988)]{sutton1988learning}
Richard~S Sutton.
\newblock Learning to predict by the methods of temporal differences.
\newblock \emph{Machine learning}, 3:\penalty0 9--44, 1988.

\bibitem[Takida et~al.(2023)Takida, Ikemiya, Shibuya, Shimada, Choi, Lai, Murata, Uesaka, Uchida, Liao, et~al.]{takida2023hq}
Yuhta Takida, Yukara Ikemiya, Takashi Shibuya, Kazuki Shimada, Woosung Choi, Chieh-Hsin Lai, Naoki Murata, Toshimitsu Uesaka, Kengo Uchida, Wei-Hsiang Liao, et~al.
\newblock Hq-vae: Hierarchical discrete representation learning with variational bayes.
\newblock \emph{arXiv preprint arXiv:2401.00365}, 2023.

\bibitem[Tian et~al.(2024)Tian, Jiang, Yuan, Peng, and Wang]{tian2024visualautoregressivemodelingscalable}
Keyu Tian, Yi Jiang, Zehuan Yuan, Bingyue Peng, and Liwei Wang.
\newblock Visual autoregressive modeling: Scalable image generation via next-scale prediction, 2024.

\bibitem[Tong et~al.(2024)Tong, Fan, Zhu, Xiong, Chen, Sinha, Rabbat, LeCun, Xie, and Liu]{tong2024metamorph}
Shengbang Tong, David Fan, Jiachen Zhu, Yunyang Xiong, Xinlei Chen, Koustuv Sinha, Michael Rabbat, Yann LeCun, Saining Xie, and Zhuang Liu.
\newblock Metamorph: Multimodal understanding and generation via instruction tuning.
\newblock \emph{arXiv preprint arXiv:2412.14164}, 2024.

\bibitem[Tschannen et~al.(2024)Tschannen, Eastwood, and Mentzer]{tschannen2024givt}
Michael Tschannen, Cian Eastwood, and Fabian Mentzer.
\newblock Givt: Generative infinite-vocabulary transformers.
\newblock In \emph{European Conference on Computer Vision}, pages 292--309. Springer, 2024.

\bibitem[Vahdat et~al.(2021)Vahdat, Kreis, and Kautz]{vahdat2021scorebasedgenerativemodelinglatent}
Arash Vahdat, Karsten Kreis, and Jan Kautz.
\newblock Score-based generative modeling in latent space, 2021.

\bibitem[Van~den Oord et~al.(2016)Van~den Oord, Kalchbrenner, Espeholt, Vinyals, Graves, et~al.]{van2016conditional}
Aaron Van~den Oord, Nal Kalchbrenner, Lasse Espeholt, Oriol Vinyals, Alex Graves, et~al.
\newblock Conditional image generation with pixelcnn decoders.
\newblock \emph{Advances in neural information processing systems}, 29, 2016.

\bibitem[Van Den~Oord et~al.(2016)Van Den~Oord, Kalchbrenner, and Kavukcuoglu]{van2016pixel}
A{\"a}ron Van Den~Oord, Nal Kalchbrenner, and Koray Kavukcuoglu.
\newblock Pixel recurrent neural networks.
\newblock In \emph{International conference on machine learning}, pages 1747--1756. PMLR, 2016.

\bibitem[Van Den~Oord et~al.(2017)Van Den~Oord, Vinyals, et~al.]{van2017neural}
Aaron Van Den~Oord, Oriol Vinyals, et~al.
\newblock Neural discrete representation learning.
\newblock \emph{Advances in neural information processing systems}, 30, 2017.

\bibitem[Vaswani et~al.(2023)Vaswani, Shazeer, Parmar, Uszkoreit, Jones, Gomez, Kaiser, and Polosukhin]{vaswani2023attentionneed}
Ashish Vaswani, Noam Shazeer, Niki Parmar, Jakob Uszkoreit, Llion Jones, Aidan~N. Gomez, Lukasz Kaiser, and Illia Polosukhin.
\newblock Attention is all you need, 2023.

\bibitem[Vincent et~al.(2008)Vincent, Larochelle, Bengio, and Manzagol]{vincent2008extracting}
Pascal Vincent, Hugo Larochelle, Yoshua Bengio, and Pierre-Antoine Manzagol.
\newblock Extracting and composing robust features with denoising autoencoders.
\newblock In \emph{Proceedings of the 25th international conference on Machine learning}, pages 1096--1103, 2008.

\bibitem[Wang et~al.(2021)Wang, Zhu, Adam, Yuille, and Chen]{wang2021maxdeeplabendtoendpanopticsegmentation}
Huiyu Wang, Yukun Zhu, Hartwig Adam, Alan Yuille, and Liang-Chieh Chen.
\newblock Max-deeplab: End-to-end panoptic segmentation with mask transformers, 2021.

\bibitem[Wang et~al.(2024)Wang, Ren, Lin, Han, Guo, Yang, Zou, Feng, and Liu]{wang2024parallelized}
Yuqing Wang, Shuhuai Ren, Zhijie Lin, Yujin Han, Haoyuan Guo, Zhenheng Yang, Difan Zou, Jiashi Feng, and Xihui Liu.
\newblock Parallelized autoregressive visual generation.
\newblock \emph{arXiv preprint arXiv:2412.15119}, 2024.

\bibitem[Wang et~al.(2004)Wang, Bovik, Sheikh, and Simoncelli]{wang2004image}
Zhou Wang, Alan~C Bovik, Hamid~R Sheikh, and Eero~P Simoncelli.
\newblock Image quality assessment: from error visibility to structural similarity.
\newblock \emph{IEEE transactions on image processing}, 13\penalty0 (4):\penalty0 600--612, 2004.

\bibitem[Weber et~al.(2024)Weber, Yu, Yu, Deng, Shen, Cremers, and Chen]{weber2024maskbit}
Mark Weber, Lijun Yu, Qihang Yu, Xueqing Deng, Xiaohui Shen, Daniel Cremers, and Liang-Chieh Chen.
\newblock Maskbit: Embedding-free image generation via bit tokens.
\newblock \emph{arXiv preprint arXiv:2409.16211}, 2024.

\bibitem[Wu et~al.(2024)Wu, Jiang, Ma, Liu, Zhao, Yuan, Bai, and Bai]{wu2024liquid}
Junfeng Wu, Yi Jiang, Chuofan Ma, Yuliang Liu, Hengshuang Zhao, Zehuan Yuan, Song Bai, and Xiang Bai.
\newblock Liquid: Language models are scalable multi-modal generators.
\newblock \emph{arXiv preprint arXiv:2412.04332}, 2024.

\bibitem[Xu et~al.()Xu, Zhang, Zhao, Li, Wang, Chen, Li, Raj, Johnson-Roberson, Scherer, et~al.]{xuscalable}
Xiaohao Xu, Tianyi Zhang, Shibo Zhao, Xiang Li, Sibo Wang, Yongqi Chen, Ye Li, Bhiksha Raj, Matthew Johnson-Roberson, Sebastian Scherer, et~al.
\newblock Scalable benchmarking and robust learning for noise-free ego-motion and 3d reconstruction from noisy video.
\newblock In \emph{The Thirteenth International Conference on Learning Representations}.

\bibitem[Yang et~al.(2024)Yang, Kang, Huang, Xu, Feng, and Zhao]{yang2024depth}
Lihe Yang, Bingyi Kang, Zilong Huang, Xiaogang Xu, Jiashi Feng, and Hengshuang Zhao.
\newblock Depth anything: Unleashing the power of large-scale unlabeled data.
\newblock In \emph{Proceedings of the IEEE/CVF Conference on Computer Vision and Pattern Recognition}, pages 10371--10381, 2024.

\bibitem[Yao and Wang(2025)]{yao2025reconstruction}
Jingfeng Yao and Xinggang Wang.
\newblock Reconstruction vs. generation: Taming optimization dilemma in latent diffusion models.
\newblock \emph{arXiv preprint arXiv:2501.01423}, 2025.

\bibitem[Yu et~al.(2023{\natexlab{a}})Yu, Cheng, Sohn, Lezama, Zhang, Chang, Hauptmann, Yang, Hao, Essa, et~al.]{yu2023magvit}
Lijun Yu, Yong Cheng, Kihyuk Sohn, Jos{\'e} Lezama, Han Zhang, Huiwen Chang, Alexander~G Hauptmann, Ming-Hsuan Yang, Yuan Hao, Irfan Essa, et~al.
\newblock Magvit: Masked generative video transformer.
\newblock In \emph{Proceedings of the IEEE/CVF Conference on Computer Vision and Pattern Recognition}, pages 10459--10469, 2023{\natexlab{a}}.

\bibitem[Yu et~al.(2023{\natexlab{b}})Yu, Lezama, Gundavarapu, Versari, Sohn, Minnen, Cheng, Gupta, Gu, Hauptmann, Gong, Yang, Essa, Ross, and Jiang]{yu2023language}
Lijun Yu, José Lezama, Nitesh~B. Gundavarapu, Luca Versari, Kihyuk Sohn, David Minnen, Yong Cheng, Agrim Gupta, Xiuye Gu, Alexander~G. Hauptmann, Boqing Gong, Ming-Hsuan Yang, Irfan Essa, David~A. Ross, and Lu Jiang.
\newblock Language model beats diffusion -- tokenizer is key to visual generation, 2023{\natexlab{b}}.

\bibitem[Yu et~al.(2024{\natexlab{a}})Yu, Cheng, Wang, Kumar, Macherey, Huang, Ross, Essa, Bisk, Yang, et~al.]{yu2024spae}
Lijun Yu, Yong Cheng, Zhiruo Wang, Vivek Kumar, Wolfgang Macherey, Yanping Huang, David Ross, Irfan Essa, Yonatan Bisk, Ming-Hsuan Yang, et~al.
\newblock Spae: Semantic pyramid autoencoder for multimodal generation with frozen llms.
\newblock \emph{Advances in Neural Information Processing Systems}, 36, 2024{\natexlab{a}}.

\bibitem[Yu et~al.(2024{\natexlab{b}})Yu, He, Deng, Shen, and Chen]{yu2024randomized}
Qihang Yu, Ju He, Xueqing Deng, Xiaohui Shen, and Liang-Chieh Chen.
\newblock Randomized autoregressive visual generation.
\newblock \emph{arXiv preprint arXiv:2411.00776}, 2024{\natexlab{b}}.

\bibitem[Yu et~al.(2024{\natexlab{c}})Yu, Weber, Deng, Shen, Cremers, and Chen]{yu2024imageworth32tokens}
Qihang Yu, Mark Weber, Xueqing Deng, Xiaohui Shen, Daniel Cremers, and Liang-Chieh Chen.
\newblock An image is worth 32 tokens for reconstruction and generation, 2024{\natexlab{c}}.

\bibitem[Zha et~al.(2024)Zha, Yu, Fathi, Ross, Schmid, Katabi, and Gu]{zha2024language}
Kaiwen Zha, Lijun Yu, Alireza Fathi, David~A Ross, Cordelia Schmid, Dina Katabi, and Xiuye Gu.
\newblock Language-guided image tokenization for generation.
\newblock \emph{arXiv preprint arXiv:2412.05796}, 2024.

\bibitem[Zhao et~al.(2024)Zhao, Xiong, and Kr{\"a}henb{\"u}hl]{zhao2024image}
Yue Zhao, Yuanjun Xiong, and Philipp Kr{\"a}henb{\"u}hl.
\newblock Image and video tokenization with binary spherical quantization.
\newblock \emph{arXiv preprint arXiv:2406.07548}, 2024.

\bibitem[Zheng et~al.(2022)Zheng, Vuong, Cai, and Phung]{zheng2022movqmodulatingquantizedvectors}
Chuanxia Zheng, Long~Tung Vuong, Jianfei Cai, and Dinh Phung.
\newblock Movq: Modulating quantized vectors for high-fidelity image generation, 2022.

\bibitem[Zhu et~al.(2024{\natexlab{a}})Zhu, Wei, Lu, and Chen]{zhu2024scaling}
Lei Zhu, Fangyun Wei, Yanye Lu, and Dong Chen.
\newblock Scaling the codebook size of vqgan to 100,000 with a utilization rate of 99\%.
\newblock \emph{arXiv preprint arXiv:2406.11837}, 2024{\natexlab{a}}.

\bibitem[Zhu et~al.(2010)Zhu, Su, Lu, Li, Wang, and Dai]{zhu2010deformable}
X Zhu, W Su, L Lu, B Li, X Wang, and J Dai.
\newblock Deformable detr: Deformable transformers for end-to-end object detection. arxiv 2020.
\newblock \emph{arXiv preprint arXiv:2010.04159}, 2010.

\bibitem[Zhu et~al.(2024{\natexlab{b}})Zhu, Li, Xin, and Xu]{zhu2024addressing}
Yongxin Zhu, Bocheng Li, Yifei Xin, and Linli Xu.
\newblock Addressing representation collapse in vector quantized models with one linear layer.
\newblock \emph{arXiv preprint arXiv:2411.02038}, 2024{\natexlab{b}}.

\bibitem[Zhu et~al.(2024{\natexlab{c}})Zhu, Li, Zhang, Li, Xu, and Bing]{zhu2024stabilize}
Yongxin Zhu, Bocheng Li, Hang Zhang, Xin Li, Linli Xu, and Lidong Bing.
\newblock Stabilize the latent space for image autoregressive modeling: A unified perspective.
\newblock \emph{arXiv preprint arXiv:2410.12490}, 2024{\natexlab{c}}.

\end{thebibliography}
}
    
\end{document}